\documentclass{article}

\usepackage{arxiv}

\usepackage[utf8]{inputenc} % allow utf-8 input
\usepackage[T1]{fontenc}    % use 8-bit T1 fonts
\usepackage{url}            % simple URL typesetting
\usepackage{booktabs}       % professional-quality tables
\usepackage{amsfonts}       % blackboard math symbols
\usepackage{nicefrac}       % compact symbols for 1/2, etc.
\usepackage{microtype}      % microtypography
\usepackage{graphicx}
\usepackage[numbers]{natbib}
\usepackage{doi}
\usepackage{multirow}
\usepackage{booktabs}
\usepackage{setspace}
\usepackage[xcolor]{changebar}
\newcommand{\CellLength}{14mm}
\usepackage{amsmath}
\usepackage{xcolor}
\usepackage{ulem}
\usepackage{changebar}
\usepackage{hyperref}       % hyperlinks

\newcommand{\deleted}[1]{%
}
\newcommand{\added}[1]{%
    \textcolor{black}{#1}% Green text
}
\newcommand{\replaced}[2]{%
    \textcolor{black}{#2}% Orange text
}

\title{Semi-Self-Supervised Domain Adaptation: Developing Deep Learning Models with Limited Annotated Data for Wheat Head Segmentation}

\author{ \href{https://orcid.org/0000-0002-1670-4489}{\includegraphics[scale=0.06]{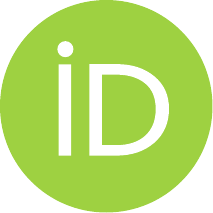}\hspace{1mm}Alireza Ghanbari}\\
	Department of Mathematics\\
     Faculty of Sciences, University of Qom\\
     Qom, Iran\\
	\And
	\href{https://orcid.org/0000-0003-2759-4606}{\includegraphics[scale=0.06]{orcid.pdf}\hspace{1mm}Gholamhassan Shirdel} \\
	Department of Mathematics\\
     Faculty of Sciences, University of Qom\\
     Qom, Iran\\
	\texttt{} \\
	\AND
    \href{https://orcid.org/0000-0002-5673-8210}
    {\includegraphics[scale=0.06]{orcid.pdf}\hspace{1mm}Farhad Maleki} \\
	Department of Computer Science\\
    University of Calgary\\
    Calgary, Alberta, Canada \\
	\texttt{farhad.maleki1@ucalgary.ca} \\
}

\renewcommand{\shorttitle}{Semi-Self-Supervised Domain Adaptation}

\begin{document}
\maketitle

\begin{abstract}
	Precision agriculture involves the application of advanced technologies to improve agricultural productivity, efficiency, and profitability while minimizing waste and environmental impact. Deep learning approaches enable automated decision-making \replaced{by analyzing visual data}{for many visual tasks}. However, \added{in the agricultural domain,} variability in growth stages and environmental conditions, such as weather and lighting, presents significant challenges to developing deep learning-based techniques that generalize across different conditions. \deleted{In addition,} \replaced{t}{T}he resource-intensive nature of creating extensive annotated datasets that capture these variabilities \replaced{also}{further} hinders the widespread adoption of these approaches.
To tackle these issues, we introduce a semi-self-supervised domain adaptation technique based on deep convolutional neural networks with a probabilistic diffusion process, requiring minimal manual data annotation. Using only three manually annotated images and a selection of video clips from wheat fields, we generated a large-scale computationally annotated dataset of image-mask pairs and a large dataset of unannotated images extracted from video frames.
We developed a two-branch convolutional encoder-decoder model architecture that uses both synthesized image-mask pairs and unannotated images, enabling effective adaptation to real images. The proposed model achieved a Dice score of 80.7\% on an internal test dataset and a Dice score of 64.8\% on an external test set\replaced{---}{,} composed of images from five countries and spanning 18 domains\replaced{---}{,} indicating its potential to develop generalizable solutions that could encourage the wider adoption of advanced technologies in agriculture.
\end{abstract}

% keywords can be removed

\section{Introduction}
Precision agriculture refers to the use of advanced technologies, including GPS guidance, control systems, sensors, robotics, drones, autonomous vehicles, and variable rate technology, to optimize farm management. It aims to reduce costs and increase yield in farming while ensuring sustainability and environmental protection~\cite{oliver2013precision}.
Deep learning (DL) methodologies can offer efficient, automated, and data-driven decision-making in agricultural practices. DL has demonstrated significant advancements in visual data analysis across various tasks, including image classification~\cite{tan2019efficientnet, najafian2023detection}, object detection~\cite{wang2023yolov7, najafian2021semi}, semantic segmentation~\cite{nazeri2021exploring, mo2022review, najafian2023semi, kirillov2023segment}, and instance segmentation~\cite{8237584, hafiz2020survey, champ2020instance}. Automated visual monitoring of agricultural fields can aid in the early identification of issues such as pest infestations, diseases, and nutrient deficiencies while optimizing resource utilization. Consequently, this approach facilitates timely interventions, which can lead to enhanced crop yields, improved quality of harvests, and increased overall operational efficiency, thereby ensuring sustainability.\par
However, the wide adoption of DL approaches for crop monitoring faces significant challenges. Agricultural fields are constantly changing environments. For example, a crop field in the early growth stages is substantially different from the same field in the later stages of growth. Additionally, these systems must operate accurately under various weather and lighting conditions.
These challenges the generalizability of DL models because models trained on data from a specific growth stage of a field might not generalize well to the same crop at different growth stages. In the DL context, this phenomenon is known as a distribution shift~\cite{Sinha_2023_WACV,hu2023codes}. One could develop a large-scale dataset encompassing various crop growth stages and environmental conditions to alleviate this issue. However, this approach presents challenges in data collection and annotation. Collecting data from various growth stages of crop fields under various weather and lighting conditions is time-consuming. Further, annotating agricultural images is particularly challenging, often requiring pixel-level annotation. These images frequently contain numerous objects of interest---e.g., wheat spikes in a wheat field---making the data annotation process laborious.\par
Domain adaptation techniques refer to the methodologies used to alleviate distribution shift and can be divided into supervised, semi-supervised,  self-supervised, and unsupervised approaches depending on whether or not annotated data is used for domain adaptation~\cite{hwang2022large,pan2020unsupervised}. Supervised domain adaptation relies on labeled data from both source and target domains. Semi-supervised domain adaptation combines labeled data from the source domain with both labeled and unlabeled data from the target domain. Finally, unsupervised domain adaptation operates solely with unlabeled data in the target domain, focusing on learning features applicable across domains.\par
In this paper, we develop a semi-self-supervised domain adaptation approach based on a probabilistic diffusion model that operates without the need for manual annotation of data from the target domain. This alleviates the need for further manual data annotation, accelerating the model development process. Self-supervised learning refers to methodologies where supervisory signals are generated computationally from the input data, enabling learning data representation and extracting informative features without the need for manual annotation~\cite{rani2023self}. This approach allows for utilizing large-scale but unannotated datasets and has recently shown substantial progress in various image-processing tasks~\cite{rani2023self, pmlr-v139-zbontar21a}.
Deep diffusion probabilistic models, as self-supervised techniques, have shown remarkable results in image-processing tasks, especially in Generative AI~\cite{ramesh2022hierarchical, rombach2022high, croitoru2023diffusion}. In the following, we provide a mathematical description of these models.
\subsection{Background}
A Markov chain is a sequence of random variables $X_1, X_2, X_3, \ldots$ that satisfies the Markov property, which states that the probability of the system transitioning to the next state depends solely on the current state of the system and not the preceding events/states. Markov property can be mathematically expressed as follows:
\begin{equation}
P(X_{n+1} = x \mid X_0 = x_0, X_1 = x_1, X_2 = x_2, \ldots, X_n = x_n) = P(X_{n+1} = x \mid X_n = x_n)
\end{equation}
In this equation, $P$  denotes a probability distribution for the state of the system. $X_i$ is a \added{vector-valued} variable representing the state at the $i^{th}$ step, and $x, x_0, x_1, x_2, \ldots, x_n$ are specific state \added{vector} values of the system.\par
Given a data point $x_0$ from the actual data distribution, we can define a Markov chain---referred to as a forward diffusion process---by iteratively adding \added{multivariate} Gaussian noise \added{vector} to $x_0$. More specifically, at state $X_{t-1}$, where the system has a specific state of $x_{t-1}$, we add a Gaussian noise\added{---characterized by a mean vector of $\mu_t$ and a covariance matrix of $\Sigma_t^2$---}to $x_{t-1}$ to transition to a state $X_t$ with a specific value of $x_t$. The probability distribution for this transition can be described as $q(X_t=x_t\mid X_{t-1}=x_{t-1})=\mathcal{N}(x_t;\mu_t=\sqrt{1-\beta_t}x_{t-1},~\Sigma_t^2=\beta_t I)$, where $\beta_t$ is an scalar; $I$ is the $m \times m$ identity matrix; and $m$ is the cardinality of $x_i$ $(0\leq i\leq T)$.
Considering the Markov chain property, we have $$q(x_{1:T}|x_0) = \prod_{t=1}^T{q(x_t\mid x_{t-1})}$$
where $q(x_{1:T}|x_0)$ is a short notation for $q(x_1,x_2,\ldots, x_T|x_0)$. A state $x_t$ can be achieved by iteratively applying the Gaussian noise to $x_0$, $t$ times.
Instead of an iterative process, this could also be achieved in a single step, utilizing a reparametrization trick, in which $x_t$ is rewritten as 
\begin{equation}
\label{Eq:Reparametrization}
x_t = \sqrt{1-\beta_t}x_{t-1} + \sqrt{\beta_t} ~\epsilon
\end{equation}
Where $\epsilon$ is an $n$-dimensional vector with standard normal distribution $\mathcal{N}(0, I)$. It can be inferred that this reparametrization preserves the distribution of $x_t$, i.e., $x_t\sim\mathcal{N}(\mu_t=\sqrt{1-\beta_t}x_{t-1},~\Sigma_t^2=\beta_t I)$. By changing the notation as ${\alpha}_t = 1-\beta_t$, Equation~\ref{Eq:Reparametrization} can be written as:
\begin{equation}
\label{Eq:ReparametrizationAlpha}
x_t = \sqrt{\alpha_t}x_{t-1} + \sqrt{1-\alpha_t} ~\epsilon
\end{equation}
As Equation~\ref{Eq:ReparametrizationAlpha} is a recursive equation, we can further expand the equation. 
\begin{align}
\label{Eq:ReparametrizationAlphaExpanded}\nonumber
x_t =& \sqrt{\alpha_t}x_{t-1} + \sqrt{1-\alpha_t} ~\epsilon\\\nonumber
    =& \sqrt{\alpha_t}\Big(\sqrt{\alpha_{t-1}}x_{t-2} + \sqrt{1-\alpha_{t-1}} ~\epsilon\Big) + \sqrt{1-\alpha_t} ~\epsilon\\
    =& \sqrt{\alpha_t\alpha_{t-1}}x_{t-2} + \sqrt{\alpha_t(1-\alpha_{t-1})} ~\epsilon + \sqrt{1-\alpha_t} ~\epsilon
\end{align}
Since $\epsilon$ has standard normal distribution $\mathcal{N}(\mu=0,~\Sigma^2=I)$, $\sqrt{\alpha_t(1 - \alpha_{t-1})} ~\epsilon$ has a normal distribution $\mathcal{N}(\mu=0,~\Sigma^2=\alpha_t(1-\alpha_{t-1}))$ and $\sqrt{1-\alpha_t}~\epsilon$ has a normal distribution $\mathcal{N}(\mu=0, ~\Sigma^2=1-\alpha_t)$. Consequently, their summation, i.e. $\sqrt{\alpha_t(1 - \alpha_{t-1})} \epsilon + \sqrt{1-\alpha_t} \epsilon$, has a normal distribution $\mathcal{N}(\mu=0,~\Sigma^2=1-\alpha_t\alpha_{t-1})$. Therefore, Equation~\ref{Eq:ReparametrizationAlphaExpanded} can be written as:
\begin{align}
\label{Eq:ReparametrizationAlphaSummarized}
x_t =& \sqrt{\alpha_t\alpha_{t-1}}x_{t-2} + \sqrt{1-\alpha_t\alpha_{t-1}} ~\epsilon
\end{align}
By expanding Equation~\ref{Eq:ReparametrizationAlphaSummarized} further, we obtain the following Equation:
\begin{align}
\label{Eq:ReparametrizationAlphaFinal}
x_t =& \sqrt{\alpha_t\alpha_{t-1}\ldots\alpha_1}x_{0} + \sqrt{1-\alpha_t\alpha_{t-1}\ldots\alpha_1} ~\epsilon
\end{align}
Defining $\overline{\alpha}_{t}=\alpha_t\alpha_{t-1}\ldots\alpha_1$, Equation~\ref{Eq:ReparametrizationAlphaFinal} can be written concisely as follow:
\begin{align}
\label{Eq:ReparametrizationAlphaBar}
x_t =& \sqrt{\overline{\alpha}_{t}}x_{0} + \sqrt{1-\overline{\alpha}_{t}} ~\epsilon
\end{align}
Given $x_0$, Equation~\ref{Eq:ReparametrizationAlphaBar} allows generating $x_t$ in a computationally efficient manner in just one step, rather than iteratively adding noise in $t$ steps. This is achieved by sampling from the Gaussian distribution $\mathcal{N}(x_t;\mu_t=\sqrt{\overline{\alpha}_{t}}x_0,~\Sigma_t^2=(1-\overline{\alpha}_{t}) I)$.
$\overline{\alpha}_{t}$ is defined using $\alpha_1, \dots \alpha_t$, where $\alpha_i = 1- \beta_i$ $(1\leq i\leq t)$. $\beta_i$ values are often defined using a linear~\cite{chen2023importance} or cosine scheduler~\cite{chen2023importance,nichol2021improved}, with the cosine schedule leading to superior results as it smoothly transitions the noise added to the data across the diffusion steps, which is essential for controlling the quality and characteristics of the generated data in diffusion models. A Cosine scheduler can be defined as follows:
$$
\beta_t = \beta_{\text{max}} - \frac{1}{2} (\beta_{\text{max}} - \beta_{\text{min}}) \left(1 + \cos\left(\frac{t}{T} \pi\right)\right)
$$
Where $\beta_t$ represents the noise level at step $t$ (higher $\beta_t$ values correspond to higher noise levels); $t=1$ and $t=T$ result in the minimum and maximum noise levels, respectively; $T$ is the total number of steps in the diffusion process; and $t$ is the current time step. \added{All experiments in this study were conducted using $\beta_{\text{min}} = 0.0001$ and $\beta_{\text{max}} = 0.02$.}.\par
As $T$ approaches infinity, $x_T$ converges to isotropic Gaussian distribution, meaning that all components of $x_t$ are normally distributed and decorrelated. Therefore, by having a procedure for learning the reverse distribution $q(x_{t-1}\mid x_t)$, we can start with sampling $x_T$ from a normal distribution and follow the reverse distribution to arrive at $x_0$. 
\added{It should be noted that in the forward diffusion process, noise is added at each step; that is, $x_t$ is derived from adding noise to $x_{t-1}$ according to a distribution of $q(x_{t} | x_{t-1})$. In the backward diffusion process, however, the aim is to reverse the forward process by removing the noise added to $x_t$ to recover $x_{t-1}$. This is achieved by a reverse transition using a distribution denoted as $q(x_{t-1} | x_t)$.}\par
In practical terms, the true reverse distribution $q(x_{t-1} | x_t)$ is unknown and intractable to compute directly, as accurately estimating it would require complex calculations involving the entire data distribution.
Instead, we approximate $q(x_{t-1} | x_t)$ using a parameterized model, a deep neural network, denoted as $p_{\theta}(x_{t-1} | x_t)$. Given that $q(x_{t-1} | x_t)$ is also Gaussian for small enough values of $\beta_t$ in the forward diffusion process, we can choose $p_{\theta}(x_{t-1} | x_t)$ to be a Gaussian distribution. In this case, the neural network is trained to parameterize the mean and variance of this Gaussian distribution.
\section{Data and Methodology}
This section provides an in-depth explanation of the data used in this research, along with detailed descriptions of the preprocessing methods, the model architecture, and the processes for model training and evaluation. %\added{The source code for this study is available through \url{https://github.com/ARGhanbari/DualBranchSSLDomainAdaptation}.}
\subsection{Data}
\begin{figure}
    \centering
    \includegraphics[width=1.0\textwidth]{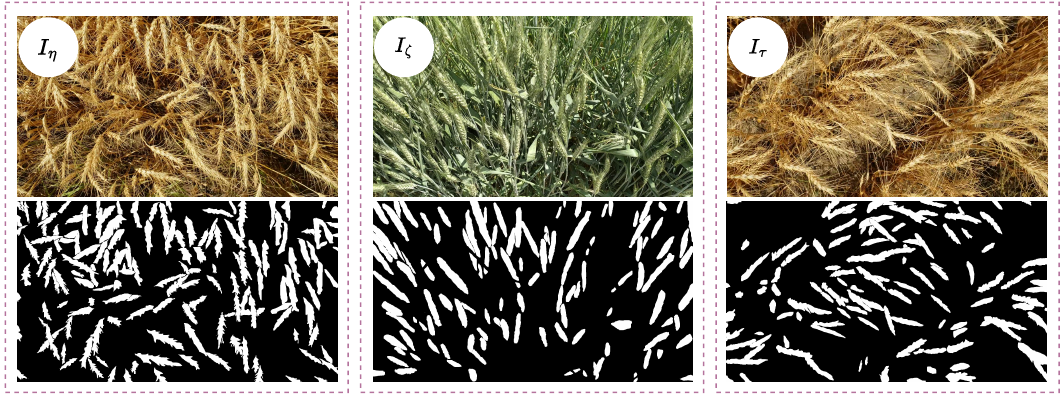}
    \caption{Three manually annotated image-mask pairs were utilized for data synthesis. We developed two training sets by synthesizing computationally annotated images using manually annotated images from the left ($I_{\eta}$) and the middle ($I_{\zeta}$), producing $8,000$ images based on $I_\eta$ and $8,000$ images based on $I_{\zeta}$. Hereafter, we refer to the $8,000$ images developed based on ($I_{\eta}$) as dataset $\mathbb{D}_{\eta}$. We refer to the set comprising the whole $16,000$ images as $\mathbb{D}_{\eta +\zeta}$. Additionally, we created a validation set by synthesizing $4,000$ images, with $2,000$ from the image on the right ($I_{\tau}$) and $2,000$ images based on $I_{\zeta}$. Hereafter, we refer to this set of $4,000$ images as $\mathbb{D}_{\zeta + \tau}$. Dataset $\mathbb{D}_{\zeta + \tau}$ was made to allow for a balanced representation of wheat field images from the early and late growth stages. All computationally annotated samples were synthesized following the methodology described by Najafian et al.~\cite{najafian2023semi}.
    }
    \label{Fig:ManuallyAnnotatedPair}
\end{figure}

\begin{figure}[!tbph]
    \centering
    \includegraphics[width=\textwidth]{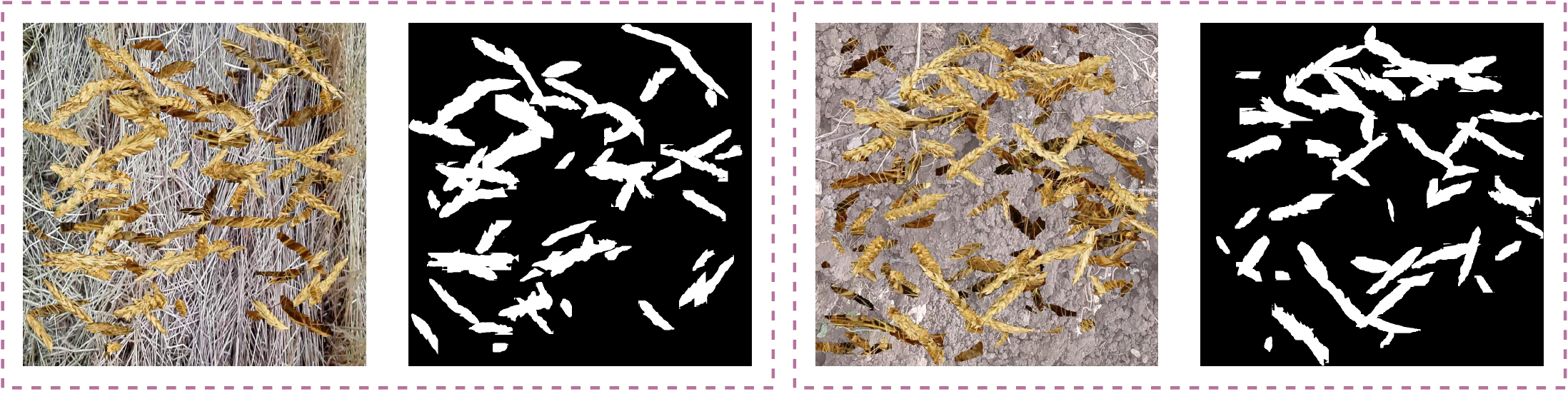}
    \caption{Examples of computationally synthesized images and their corresponding segmentation masks.}
    \label{Fig:SynthesizedImages}
\end{figure}

Figure~\ref{Fig:ManuallyAnnotatedPair} illustrates the three manually annotated images ($I_{\eta}$, $I_{\zeta}$, and $I_{\tau}$) used for synthesizing computationally annotated datasets.
Utilizing the methodology presented in our previous work~\cite{najafian2023semi}, we computationally synthesize three datasets: $\mathbb{D}_{\eta}$ a set of $8,000$ images drived from $I_{\eta}$, $\mathbb{D}_{\eta +\zeta}$  a set of $16,000$ images derived from $I_{\eta}$ and $I_{\zeta}$, and $\mathbb{D}_{\zeta + \tau}$ a set of $4,000$ images derived from $I_{\zeta}$ and $I_{\tau}$.
Figure~\ref{Fig:SynthesizedImages} illustrates examples of synthesized images and their corresponding segmentation masks.
In addition to these manually and computationally annotated images, for the training process, we utilize $10,592$ unannotated image frames extracted from two video clips of wheat fields, with $5,296$ frames from each. \added{Hereafter, we refer to these datasets as $\mathbb{D}_{\rho_1}$ and $\mathbb{D}_{\rho_2}$. We refer to the dataset resulting from combining $\mathbb{D}_{\rho_1}$ and $\mathbb{D}_{\rho_2}$ as $\mathbb{D}_{\rho}$.}\par
We also use two sets of $\Psi$ and $\Gamma$ as the internal and external test sets, as introduced by Najafian et al.~\cite{najafian2023semi}.
The set $\Psi$ comprises 100 image frames, which are randomly selected from a video clip of a wheat field and have been manually annotated to serve as an internal test set. The set $\Gamma$ represents a subset of 365 manually annotated images from the GWHD dataset~\cite{david2021global}, which includes images from five countries and spans 18 domains, covering various growth stages of wheat.\par
\subsection{Model Architecture}
\begin{figure}
    \centering
    \includegraphics[width=\linewidth]{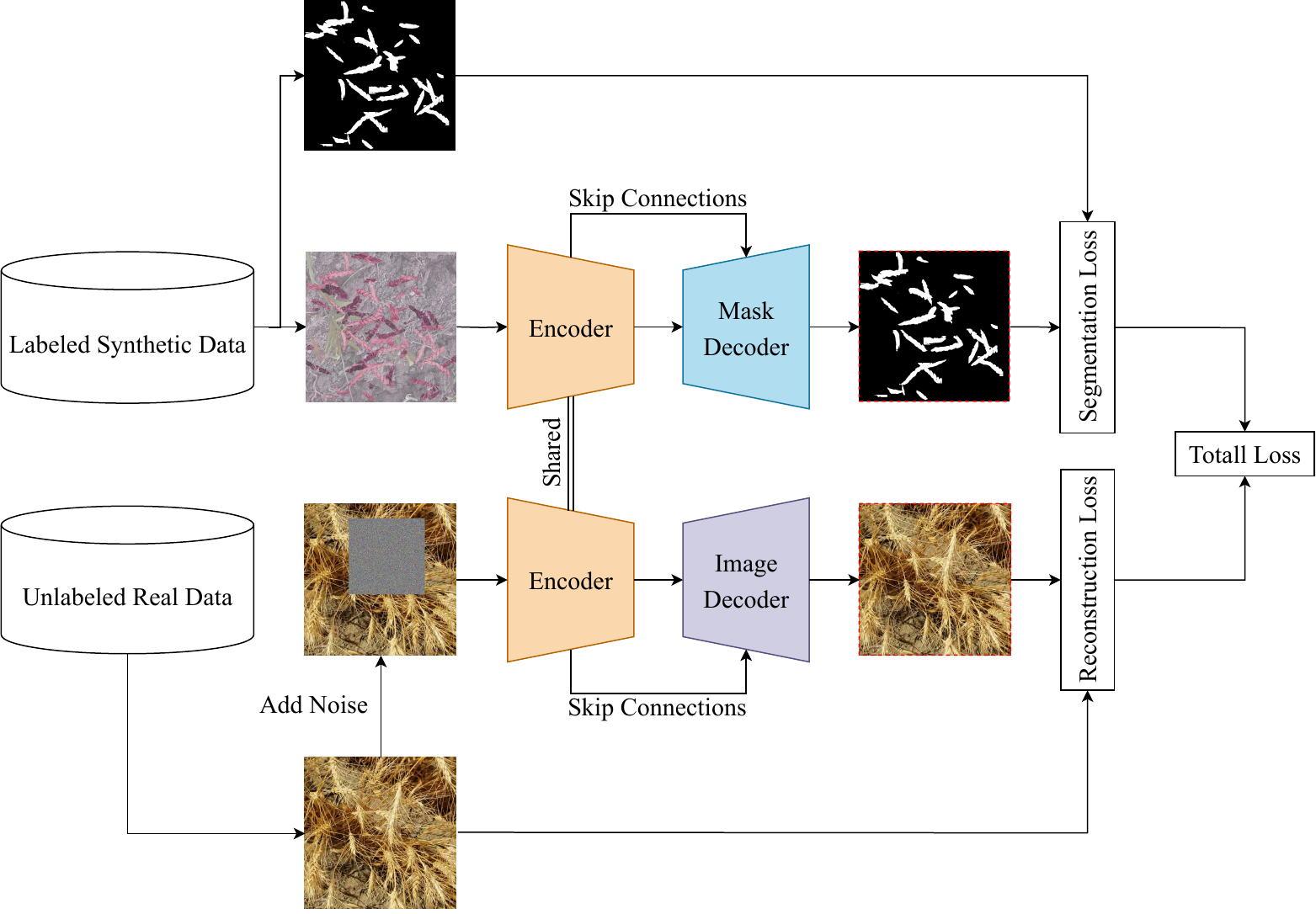}
    \caption{Schematic Representation of the Model Architecture. The encoder focuses on developing a joint image representation for both synthesized and real images, while the mask decoder aims at generating segmentation masks, and the image decoder aims at reconstructing the real images, forcing the encoder to adapt to the real images.}
    \label{Fig:ModelArchitecture}
\end{figure}

Figure~\ref{Fig:ModelArchitecture} illustrates the convolutional neural network (CNN)-based model architecture employed in this research. This architecture is designed to facilitate representation learning through a diffusion process, in addition to performing the primary task of segmentation. The segmentation task is executed using computationally synthesized images along with their corresponding masks. Meanwhile, the representation learning aspect enables adaptation to real images using solely unannotated data, thereby reducing the reliance on manual annotation.\par
We use an encoder to develop a joint representation of both synthesized and real images. Additionally, the model architecture comprises an image decoder and a mask decoder. The former is designed to reconstruct the image from the output of the encoder, thereby enforcing learning features from real images. On the other hand, the mask \replaced{encoder}{decoder} is tasked with developing a segmentation mask given the output of the encoder.\par
\begin{figure}[!tbph]
    \centering
    \includegraphics[width=0.7\textwidth]{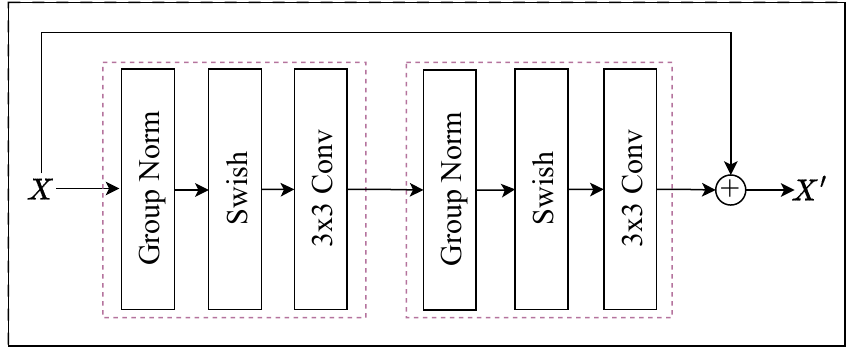}
    \caption{A ResNet block comprises three groups of operations, including convolution, GroupNorm layers, and the Swish activation function for nonlinearity. It also incorporates skip connections to enhance feature propagation.
    }
    \label{Fig:ResidualBlock}
\end{figure}

In implementing the encoder and decoders, we leveraged the Residual building blocks~\cite{he2016deep}, each comprising two pairs of convolution and GroupNorm layers, supported by the Swish activation function~\cite{ramachandran2017searching} and integrated skip connections. Figure~\ref{Fig:ResidualBlock} shows a residual building block. \par
\begin{figure}
    \centering
    \includegraphics[width=\textwidth]{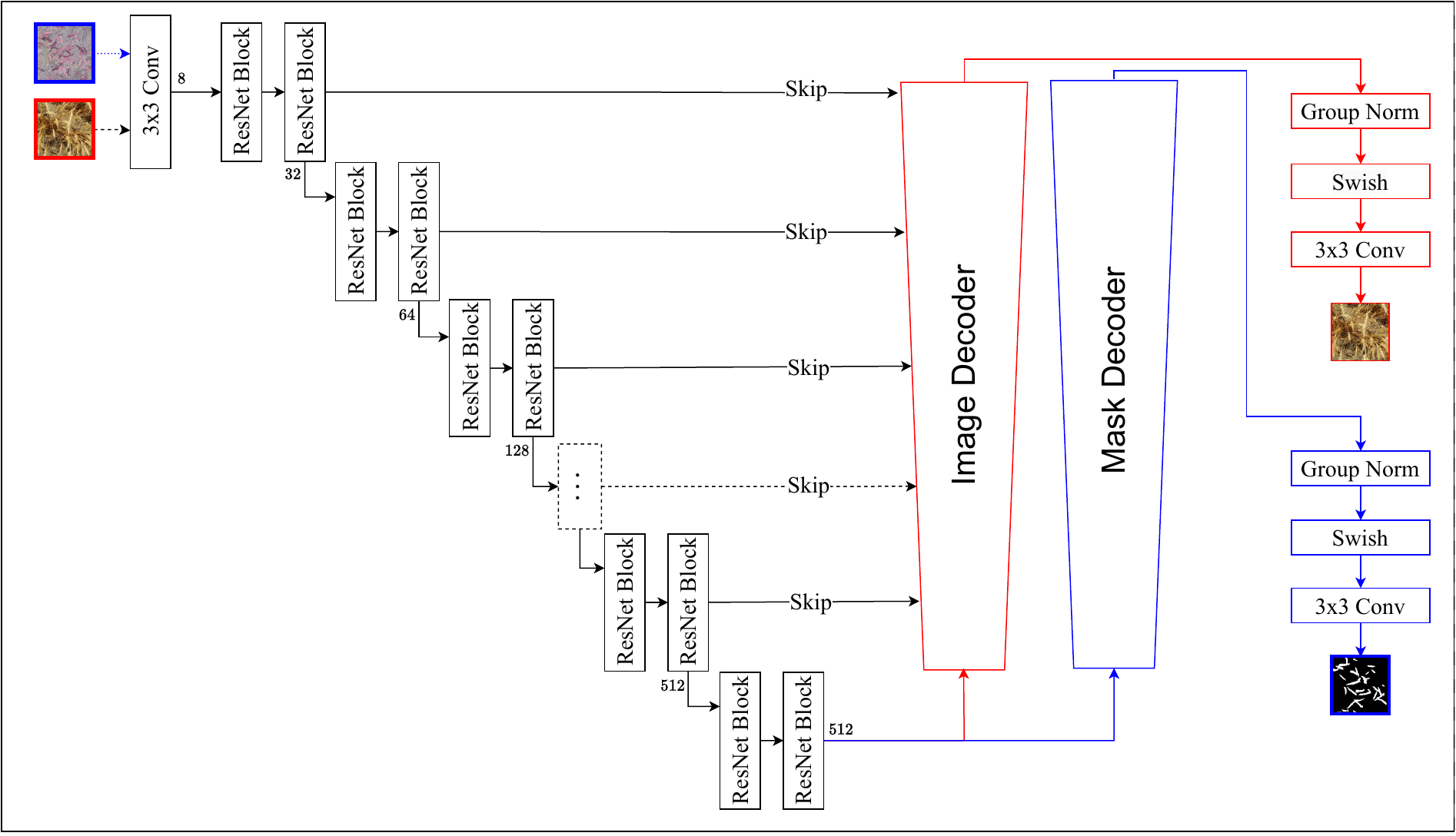}
    \caption{Encoder model architecture is designed by combining convolutional layers, ResNet blocks, and GroupNorm layers. Also, in each of the two decoding streams, we utilize concatenation instead of addition.
    }
    \label{Fig:EncoderArchitecture}
\end{figure}

Figure~\ref{Fig:EncoderArchitecture} illustrates the encoder, consisting of one convolutional layer followed by 12 ResNet building blocks organized in six levels. Additionally, two skip-connection-free ResNet blocks serve as the network bottleneck. Moreover, down-sampling operations are performed after each pair of ResNet blocks using a convolution layer with a kernel size of three and a stride of two. The Swish activation function~\cite{ramachandran2017searching} is also consistently applied as the non-linearity across the encoder layers. \par
Synthetic images undergo a series of image augmentations~\cite{buslaev2020albumentations}; then they are fed into the encoder. The outputs of the encoder are subsequently passed to a mask decoder to generate accurate masks corresponding to the input images. To calculate segmentation error, we use a linear combination of Binary Cross Entropy (BCE) loss~\cite{wazir2022histoseg} and Dice loss~\cite{wazir2022histoseg} (see Equation~\ref{Eq:SegLoss}).\par
\begin{gather}
\mathcal{L}_{BCE} = -\frac{1}{B\times N} \sum_{j=1}^{B}\sum_{i=1}^{N} \left( y_i \log(\hat{y}_i) + (1 - y_i) \log(1 - \hat{y}_i) \right) \nonumber\\
\mathcal{L}_{\text{Dice}} = 1 - \frac{2}{B} \sum_{j=1}^{B} \left( \frac{|M^{s}_j \cap \hat{M}^{s}_j|}{|M^{s}_j| + |\hat{M}^{s}_j|} \right) \nonumber\\
\mathcal{L}_{Seg} = {\lambda}_{BCE}\times \mathcal{L}_{BCE} + {\lambda}_{Dice} \times \mathcal{L}_{Dice}
\label{Eq:SegLoss}
\end{gather}
where $y_i$ is the true label of the pixel $i$, with a value of $1$ if the pixel belongs to a wheat head and $0$ otherwise. $\hat{y}_i$ represents the model prediction for the pixel $i$ being from a wheat head. $N$ is the number of pixels per image, and $B$ is the batch size. $M^{s}_j$ is the ground truth mask for the synthetic image $I^{s}_j$, and $\hat{M}^{s}_j$ is the model prediction for $I^{s}_j$.\par
We also generate noise-augmented images by applying Gaussian noise to the real image $I_j$ over $t$ $(1 \leq t \leq T)$ timesteps, as detailed in Equation~\ref{Eq:ReparametrizationAlphaBar}. These noise-augmented images are subsequently routed through the encoder, followed by the image decoder, aiming to reconstruct the original images. The error for this branch is calculated by comparing the original image with the reconstructed image using a linear combination of MSE~\cite{beheshti2010mean}, SSIM~\cite{hore2010image}, and a perceptual loss function~\cite{johnson2016perceptual}. The perceptual loss function calculates the mean absolute error between the pretrained ResNet18 output feature maps for the original image $I_j$, and the reconstructed image $\hat{I}_j$. We used the $ResNet18\_Weights.IMAGENET1K\_V1$ weights from the Torchvision Python package~\cite{NEURIPS2019_9015}. We used the feature maps generated by the convolutional layer before the final average pooling layer of the ResNet18 model.\par 
\begin{gather}
\mathcal{L}_{\text{MSE}} = \frac{1}{B \times N} \sum_{j=1}^{B}\sum_{i=1}^{N} (x_{ji} - \hat{x}_{ji})^2 \nonumber\\
\mathcal{L}_{\text{SSIM}}(I_j, \hat{I}_j) = 1 - \frac{(2\mu_{I_j} \mu_{\hat{I}_j} + c_1)(2\sigma_{I_j\hat{I}_j} + c_2)}{(\mu_{I_j}^2 + \mu_{\hat{I}_j}^2 + c_1)(\sigma_{I_j}^2 +~\sigma_{\hat{I}_j}^2 + c_2)}\nonumber\\
\mathcal{L}_{\text{Perceptual}} = \frac{1}{B} \sum_{j=1}^{B} \left\| \theta(I_j) - \theta(\hat{I}_j) \right\|_2^2\nonumber\\
\mathcal{L}_{Rec} = {\lambda}_{MSE} \times \mathcal{L}_{MSE} + {\lambda}_{SSIM} \times \mathcal{L}_{SSIM} + {\lambda}_{Perceptual} \times \mathcal{L}_{Perceptual}
\label{Eq:RecLoss}
\end{gather}
where $B$ is batch size; $N$ is the number of pixels in an image $I_j$; $x_{ji}$ is the actual value of the $i$-th pixel of $I_j$; $\hat{x}_{ji}$ is the predicted value for the $i$-th pixel of $I_j$; $\mu_{I_j}$ and $\mu_{\hat{I}_j}$ are the mean intensity values of the images $I_j$ and $\hat{I}_j$, respectively; $\sigma_{I_j}^2$ and $\sigma_{\hat{I}_j}^2$ are the variances of the images $I_j$ and $\hat{I}_j$, respectively; $\sigma_{I_j\hat{I}_j}$ is the covariance between images $I_j$ and $\hat{I}_j$; $c_1$ and $c_2$ are constants introduced to stabilize the division.
$\theta(I_j)$ and $\theta(\hat{I}_j)$ are ResNet18 features extracted from images $I_j$ and $\hat{I_j}$, respectively.\par
These losses focus on variations between the reconstructed image and the original noise-free image from different perspectives: MSE loss focuses on pixel-level variations, SSIM loss focuses on local variations, and Perceptual loss focuses on high-level global variations. The reconstruction loss is calculated as a linear combination of these loss functions, as described in Equation~\ref{Eq:RecLoss}, where $\lambda_{MSE}$, $\lambda_{SSIM}$, and $\lambda_{Perceptual}$ are constant values and model hyperparameters.\par
\subsection{Model training and evaluation}

Considering the size of the datasets, we train our models over a total of 50 training epochs. For each epoch, we iterate over all the data in the datasets. At each iteration, we select a batch of 32 synthesized image-mask pairs and a separate batch of 32 real images from our datasets. The image-mask pairs are fed into the segmentation branch of our model, which consists of an encoder and a mask decoder. The segmentation loss is then calculated according to Equation~\ref{Eq:SegLoss}. Similarly, the real images are fed into the reconstruction branch of our model, comprising the encoder (shared with the segmentation branch) and an image decoder, where the reconstruction loss is calculated as per Equation~\ref{Eq:RecLoss}. The total loss is computed by summing the segmentation loss and the reconstruction loss. Figure~\ref{Fig:ModelArchitecture} illustrates this process.\par
This dual-stream approach ensures that the encoder effectively attends to both types of data, with each decoder specializing in a task corresponding to its branch---namely, segmentation or reconstruction. Model updates are orchestrated using the AdamW optimizer~\cite{loshchilov2017decoupled}, with a learning rate set at $0.0001$. For all experiments, the coefficients ${\lambda}_{MSE}$, ${\lambda}_{SSIM}$, ${\lambda}_{Perceptual}$, ${\lambda}_{BCE}$, and ${\lambda}_{Dice}$ are set to $1$. After training for $50$ epochs, we selected the model with the highest Dice score as the best model. During the evaluation process, we assessed the model performance using the Dice score and IoU. As a baseline for comparison, we used a model developed by~\cite{najafian2023semi}.\par
\section{Results}
Table~\ref{tab:general_models_performance} presents a quantitative evaluation of the performance of our developed models, namely, model $\mathcal{F}_{\eta + {\rho}_1}$, trained on $\mathbb{D}_{\eta}$ and $\mathbb{D}_{{\rho}_1}$\deleted{combined}, and model $\mathcal{F}_{\eta + \zeta + \rho}$ trained on $\mathbb{D}_{\eta + \zeta}$ and $\mathbb{D}_{\rho}$\deleted{combined}. These models were trained using both synthesized image-mask pairs and unannotated real images extracted from video clips \added{of wheat fields}. The performance of the baseline model $\mathcal{S}$ from Najafian et al.~\cite{najafian2023semi} is also reported in Table~\ref{tab:general_models_performance}. The results demonstrate that the developed models, $\mathcal{F}_{\eta + {\rho}_1}$ and $\mathcal{F}_{\eta + \zeta + \rho}$, consistently outperformed the baseline model $\mathcal{S}$. These models rely on computationally annotated synthetic images \added{and real unannotated images extracted from the video frames of wheat fields}.\par
\begin{figure}[!h]
    \centering
    \includegraphics[width=1.0\textwidth]{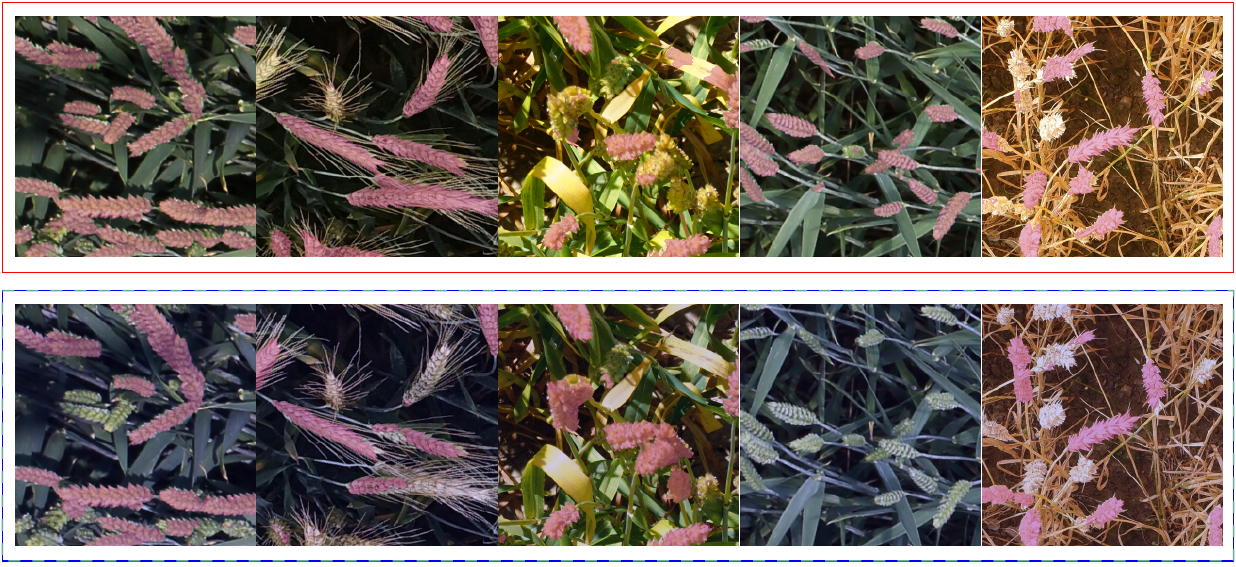}
    \caption{Showcasing the prediction performance of model $\mathcal{F}_{\eta + \zeta + \rho}$ (highlighted in a red box in the upper row) in comparison with the results obtained by model $S$~\cite{najafian2023semi} (highlighted in a blue box in the lower row) on samples from the Global Wheat Head Detection dataset~\cite{david2021global}.
    }
    \label{fig:model_prediction_performance}
\end{figure}

\begin{table}[!th]
\centering
\caption{The performance of the trained models was evaluated on our internal and external test sets using the IoU and Dice scores. Model $\mathcal{F}_{\eta + {\rho}_1}$ \deleted{represents the model}\added{was} trained on $\mathbb{D}_{\eta}$ and $\mathbb{D}_{\rho_1}$ \deleted{combined} and model $\mathcal{F}_{\eta + \zeta + \rho}$ is the result of fine-tuning model $\mathcal{F}_{\eta + {\rho}_1}$ on datasets $\mathbb{D}_{\eta + \zeta}$ and $\mathbb{D}_{\rho}$\deleted{ combined}. We also compared the performance of these two models with the model developed in~\cite{najafian2023semi}. All of these models rely exclusively on synthesized image-mask pairs and/or unannotated images.
}
\label{tab:general_models_performance}
\begin{tabular}{cccc}
\toprule
\textbf{Model}                                    & \textbf{Evaluation Method}                         & \textbf{Dice} & \textbf{IoU} \\ \toprule
$\mathcal{S}$~\cite{najafian2023semi}             & Internal test set $\Psi$                  & 0.709         & 0.565        \\
$\mathcal{F}_{\eta + \rho_1}$                     & Internal test set $\Psi$                  & 0.773         & 0.638        \\
$\mathcal{F}_{\eta + \zeta + \rho}$               & Internal test set $\Psi$                  & 0.807         & 0.686        \\ \midrule
$S$~\cite{najafian2023semi}                       & External test set $\Gamma$                & 0.367         & 0.274        \\
$\mathcal{F}_{\eta + \rho_1}$                     & External test set $\Gamma$                & 0.551         & 0.427        \\
$\mathcal{F}_{\eta + \zeta + \rho}$               & External test set $\Gamma$                & 0.648         & 0.526        \\ \bottomrule
\end{tabular}
\end{table}
Figure~\ref{fig:model_prediction_performance} presents a qualitative evaluation of the performance of models $\mathcal{F}_{\eta + \zeta + \rho}$, identified as the best-performing model, and model $\mathcal{S}$ on several different domains. In general, model $\mathcal{F}_{\eta + \zeta + \rho}$ demonstrates superior performance compared to model $\mathcal{S}$. We also reported a case, as shown in the middle column, where model $\mathcal{S}$ performs better.\par
Table~\ref{tab:per_domain_performance} presents the performance of the proposed models across various domains of the GWHD dataset. The results indicate that the proposed models generally outperform model $\mathcal{S}$ in most domains, exhibiting lower variance and greater stability.

\begin{table}[!hp]
\caption{The performance of the models on each of the 18 domains of the GWHD dataset. Model $S$\added{, which} was trained using the synthesized dataset, was reported from \deleted{ in}~\cite{najafian2023semi}. Model $\mathcal{F}_{\eta + {\rho}_1}$ \added{ was developed using dataset $\mathbb{D}_{\eta}$---consisting of 8,000 computationally annotated synthesized images---and dataset $\mathbb{D}_{\rho_1}$, which includes 5,296 unannotated real images.} Model $\mathcal{F}_{\eta + \zeta + \rho}$ was developed using \added{ dataset $\mathbb{D}_{\eta+\zeta}$---consisting of 16,000 computationally annotated synthesized images---and dataset $\mathbb{D}_{\rho}$, which includes 10,592 unannotated real images.} \deleted{two simulated datasets, each $\mathbb{D}_{\eta}$ of size 8000, and developed using a single manually annotated image frame.} We trained model $\mathcal{F}_{\eta + {\rho}_1}$ from scratch, while \deleted{further fine-tuned} model $\mathcal{F}_{\eta + \zeta + \rho}$ \added{was resulted from fine-tuning of model $\mathcal{F}_{\eta + {\rho}_1}$ \deleted{on top of $\mathcal{F}_{\eta + \rho}$} using dataset $\mathbb{D}_{\eta + \zeta}$ and dataset $\mathbb{D}_{\rho}$.} \deleted{Model $\mathcal{F}_{\eta + \zeta + \rho}$ was derived by fine-tuning model $\mathcal{F}_{\eta + \rho}$.}
}
\label{tab:per_domain_performance}
\scriptsize
\renewcommand{\arraystretch}{1.7}
\begin{minipage}[b]{0.5\linewidth} 
%\raggedleft
\centering
\begin{tabular}[5pt]{cccc}
Domain   & Model  & Dice Score \\ \hline\hline
\multirow{3}{*}{\includegraphics[width=\CellLength, height=\CellLength]{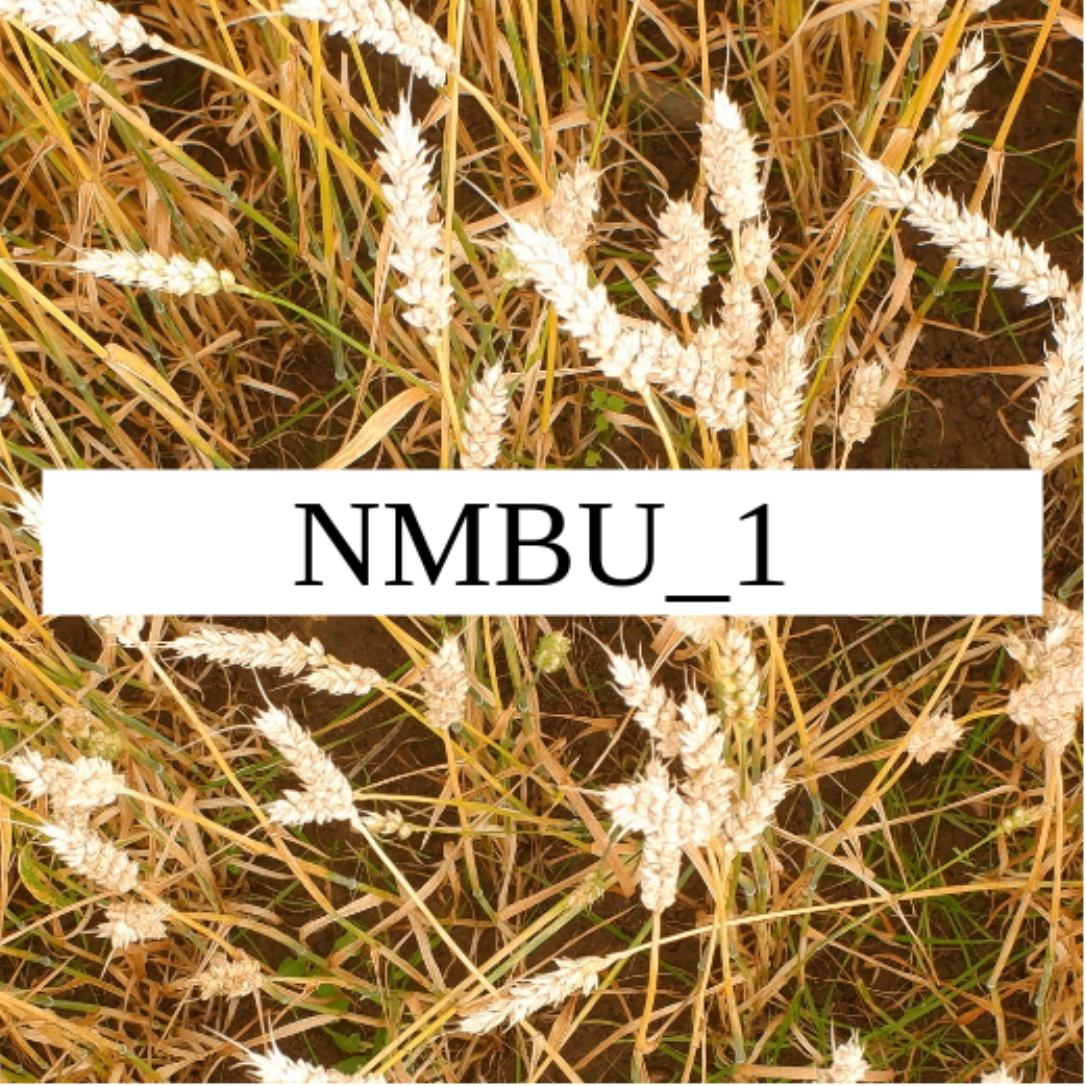}}& $\mathcal{S}$                                                                    & 0.731              \\
                  & $\mathcal{F}_{\eta + \rho_1}$                                                              & 0.660     \\
                  & $\mathcal{F}_{\eta + \zeta + \rho}$                                                           & 0.692              \\ \hline
\multirow{3}{*}{\includegraphics[width=\CellLength, height=\CellLength]{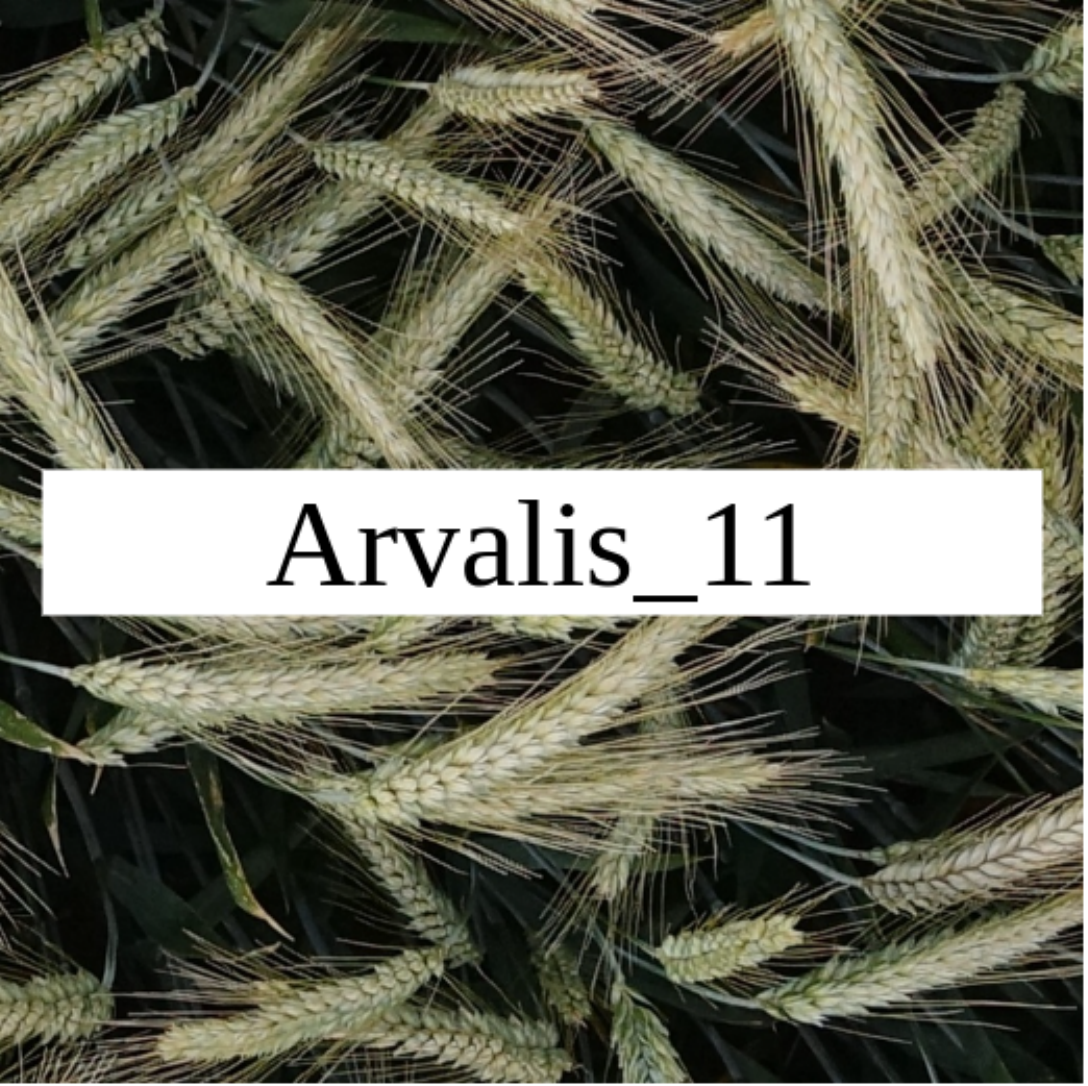}} & $\mathcal{S}$                                                                    & 0.848              \\
                  & $\mathcal{F}_{\eta + \rho_1}$                                                             & 0.815     \\
                  & $\mathcal{F}_{\eta + \zeta + \rho}$                                                            & 0.857              \\ \hline
\multirow{3}{*}{\includegraphics[width=\CellLength, height=\CellLength]{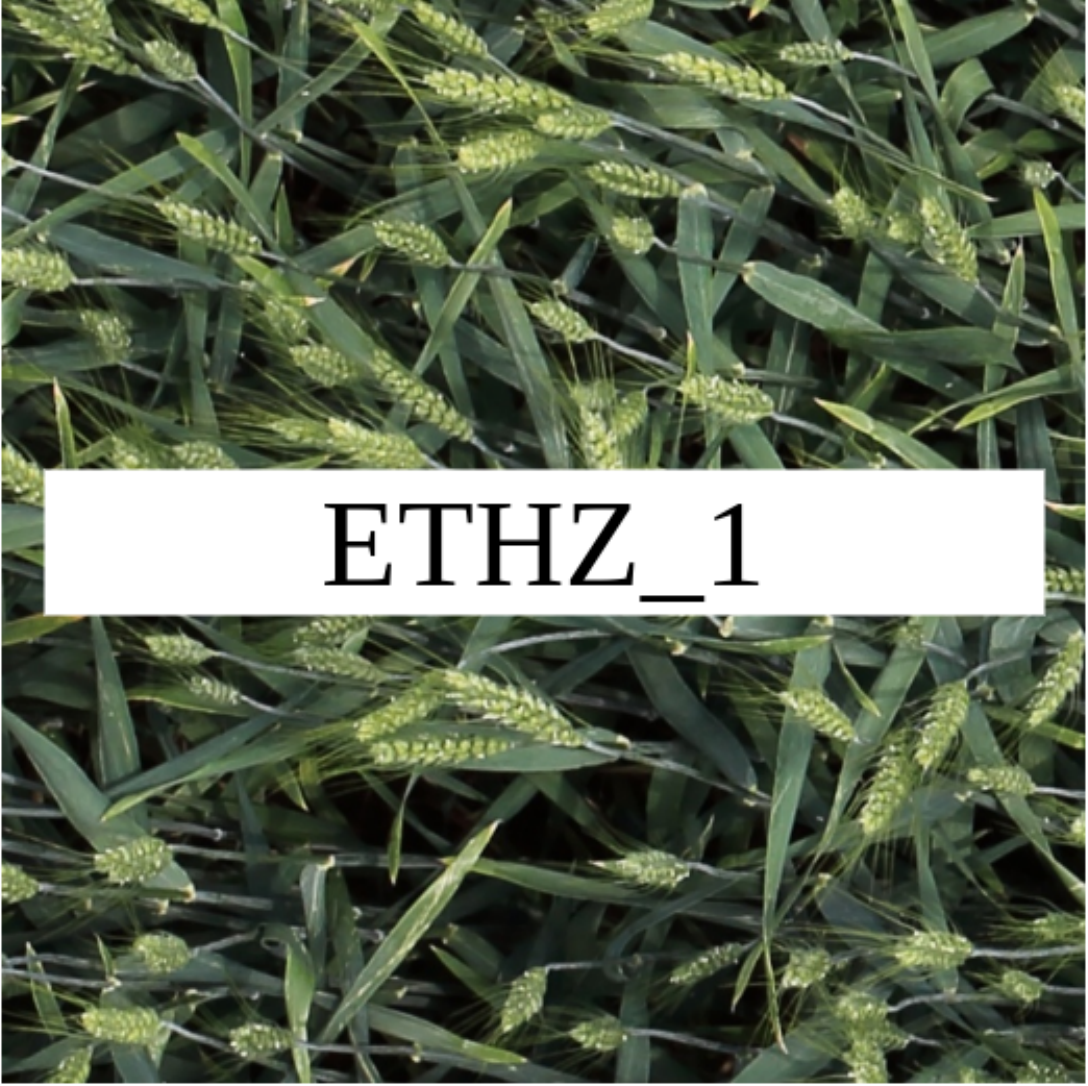}} & $\mathcal{S}$                                                                    & 0.309              \\
                  & $\mathcal{F}_{\eta + \rho_1}$                                                              & 0.764     \\
                  & $\mathcal{F}_{\eta + \zeta + \rho}$                                                           & 0.812              \\ \hline
\multirow{3}{*}{\includegraphics[width=\CellLength, height=\CellLength]{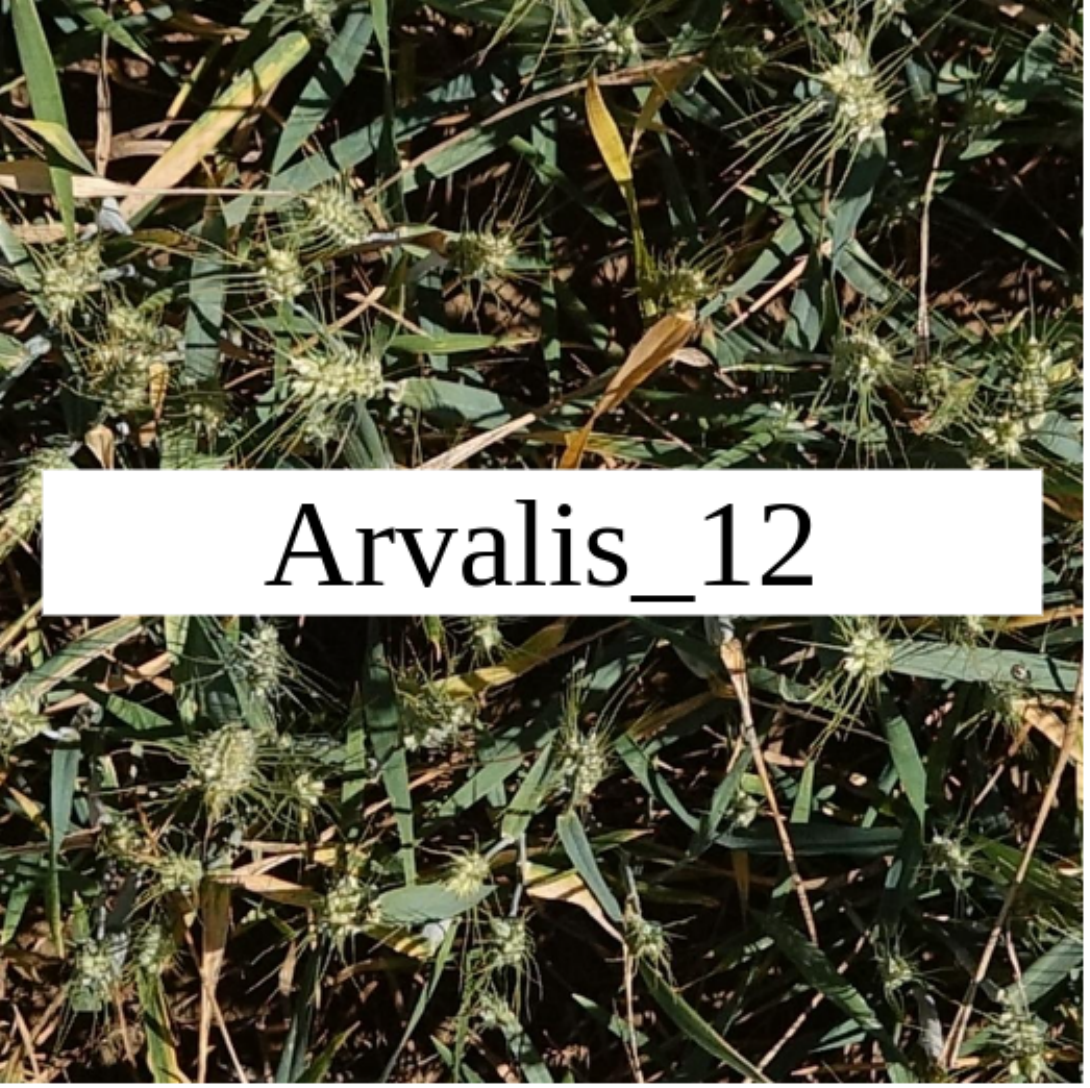}} & $\mathcal{S}$                                                                    & 0.240              \\
                  & $\mathcal{F}_{\eta + \rho_1}$                                                             & 0.503     \\
                  & $\mathcal{F}_{\eta + \zeta + \rho}$                                                            & 0.431              \\ \hline
\multirow{3}{*}{\includegraphics[width=\CellLength, height=\CellLength]{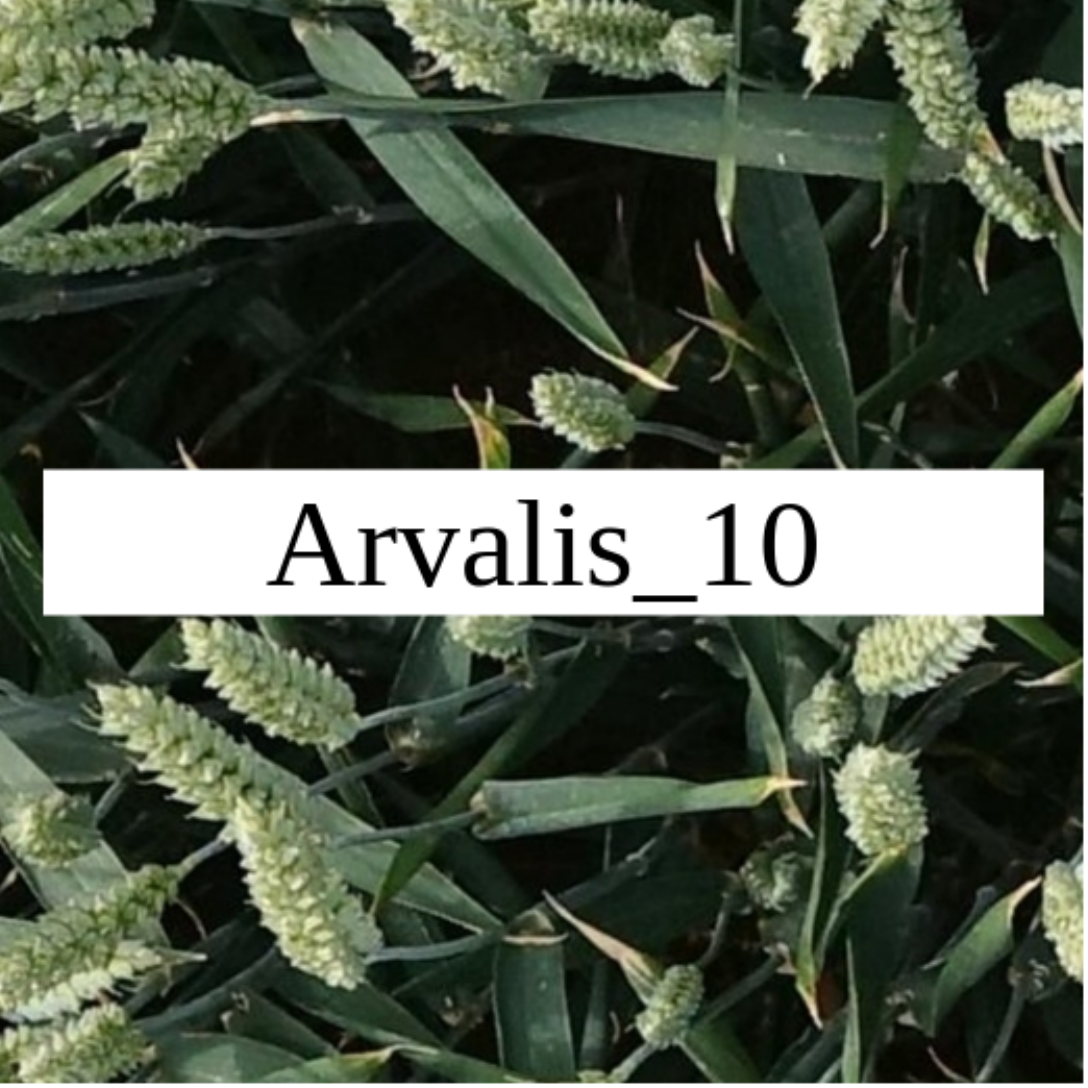}} & $\mathcal{S}$                                                                    & 0.794              \\
                  & $\mathcal{F}_{\eta + \rho_1}$                                                            & 0.715     \\
                  & $\mathcal{F}_{\eta + \zeta + \rho}$                                                            & 0.698              \\ \hline
\multirow{3}{*}{\includegraphics[width=\CellLength, height=\CellLength]{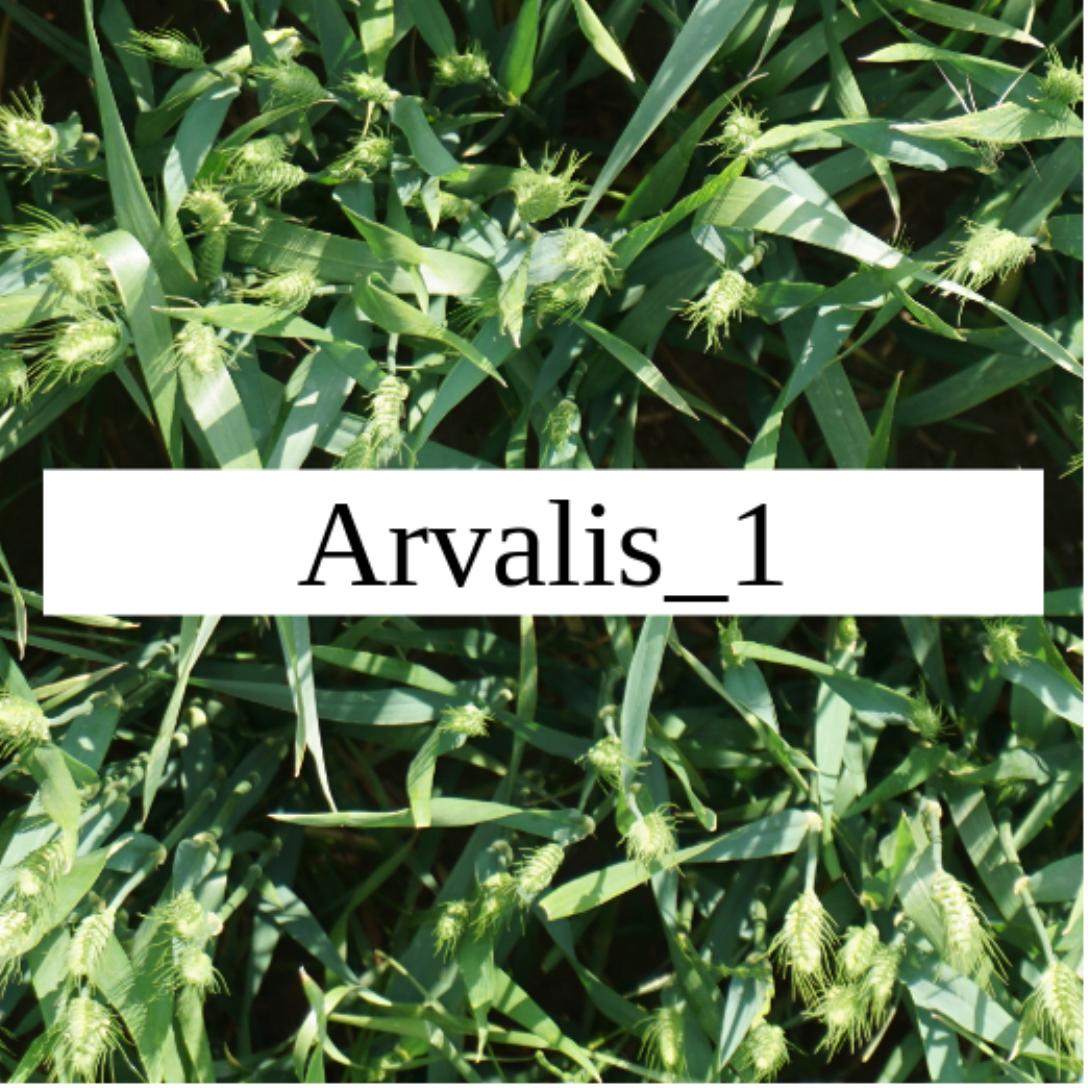}} & $\mathcal{S}$                                                                   & 0.389              \\
                  & $\mathcal{F}_{\eta + \rho_1}$                                                             & 0.479              \\
                  & $\mathcal{F}_{\eta + \zeta + \rho}$                                                          & 0.605              \\ \hline
\multirow{3}{*}{\includegraphics[width=\CellLength, height=\CellLength]{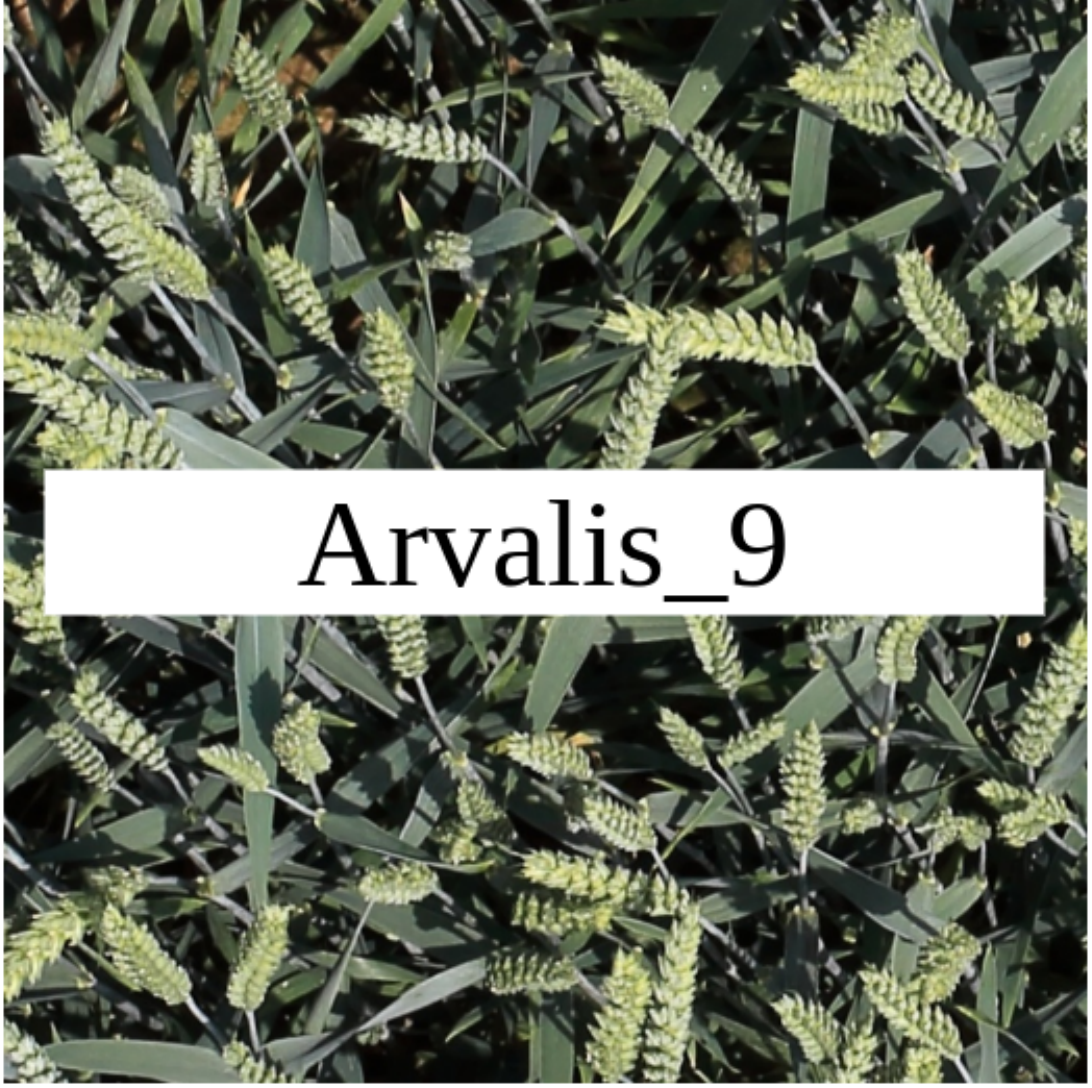}} & $\mathcal{S}$                                                                   & 0.509              \\
                  & $\mathcal{F}_{\eta + \rho_1}$                                                              & 0.655              \\
                  & $\mathcal{F}_{\eta + \zeta + \rho}$                                                           & 0.625              \\ \hline
\multirow{3}{*}{\includegraphics[width=\CellLength, height=\CellLength]{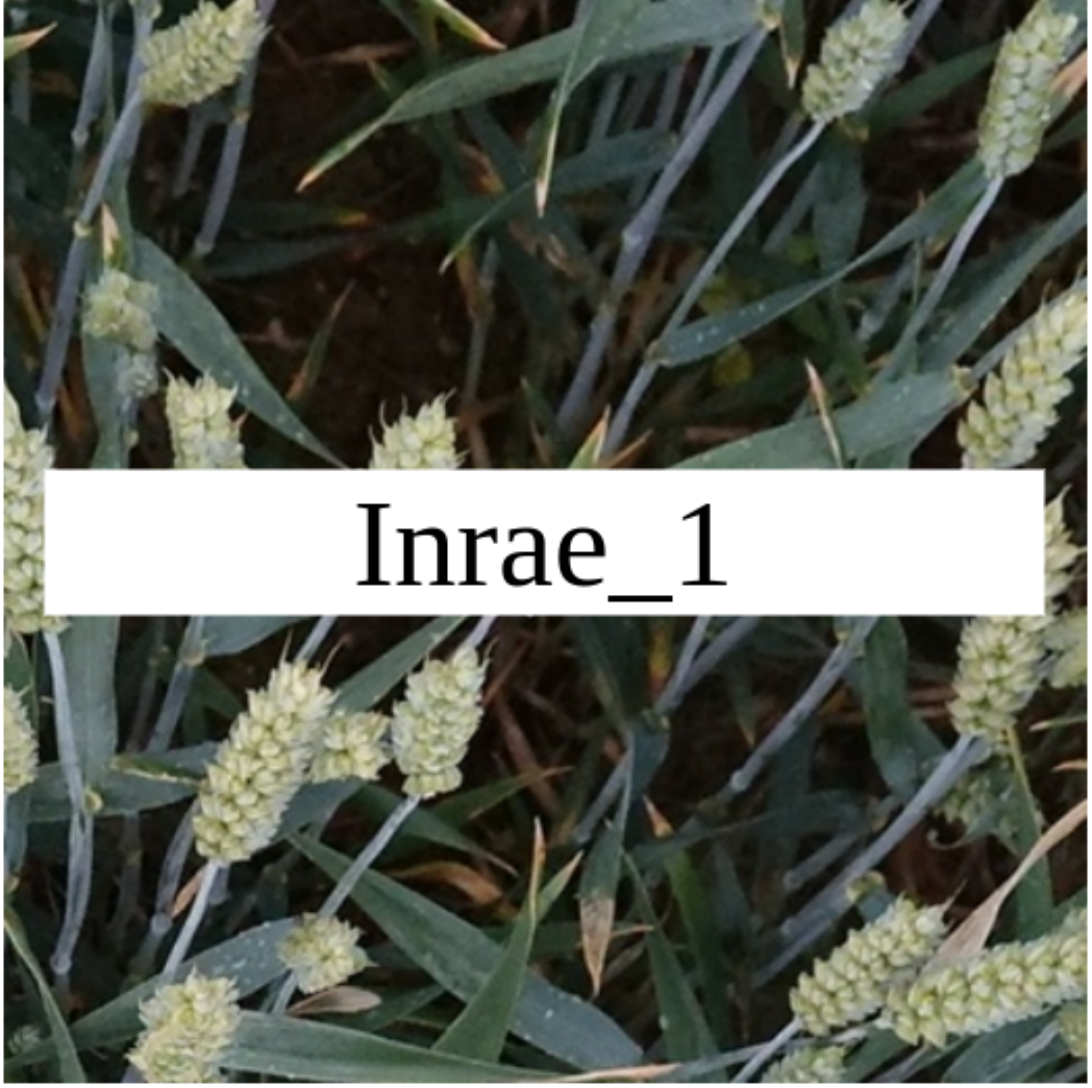}} & $\mathcal{S}$                                                                    & 0.859              \\
                  & $\mathcal{F}_{\eta + \rho_1}$                                                             & 0.674     \\
                  & $\mathcal{F}_{\eta + \zeta + \rho}$                                                           & 0.671              \\ \hline
\multirow{3}{*}{\includegraphics[width=\CellLength, height=\CellLength]{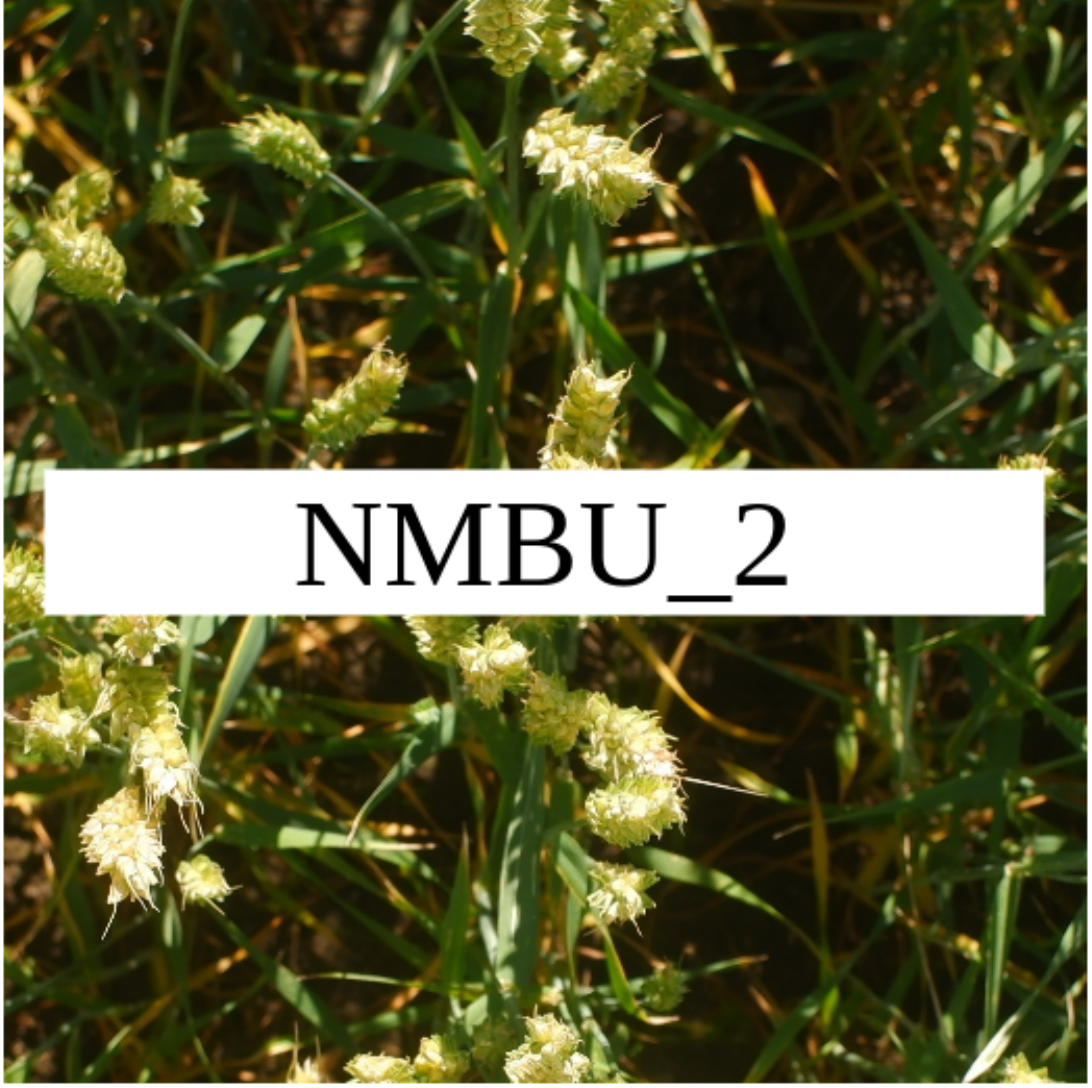}} & $\mathcal{S}$                                                                   & 0.539              \\
                  & $\mathcal{F}_{\eta + \rho_1}$                                                            & 0.586     \\
                  & $\mathcal{F}_{\eta + \zeta + \rho}$                                                           & 0.579              \\ \hline\hline
\end{tabular}
\end{minipage}%
% Second Minipage. 
\begin{minipage}[b]{0.5\linewidth}
%\raggedright
%\caption{Prices for 2011}
%\label{tab:prices2011}
\centering
\begin{tabular}{ccc}
Domain   & Model & Dice Score \\ \hline\hline
\multirow{3}{*}{\includegraphics[width=\CellLength, height=\CellLength]{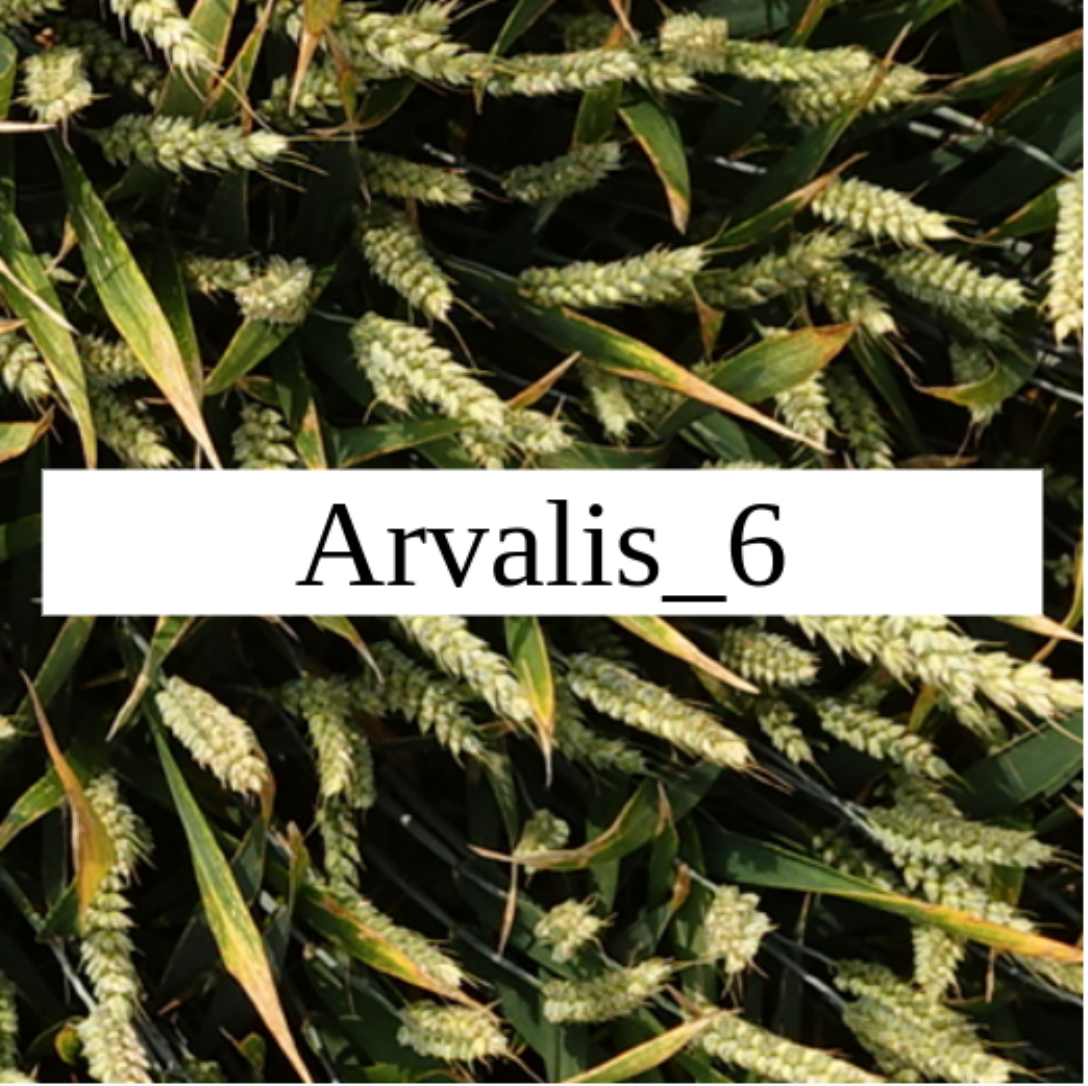}} & $\mathcal{S}$                                                                    & 0.711              \\
                  & $\mathcal{F}_{\eta + \rho_1}$                                                            & 0.644     \\
                  & $\mathcal{F}_{\eta + \zeta + \rho}$                                                            & 0.759              \\ \hline
\multirow{3}{*}{\includegraphics[width=\CellLength, height=\CellLength]{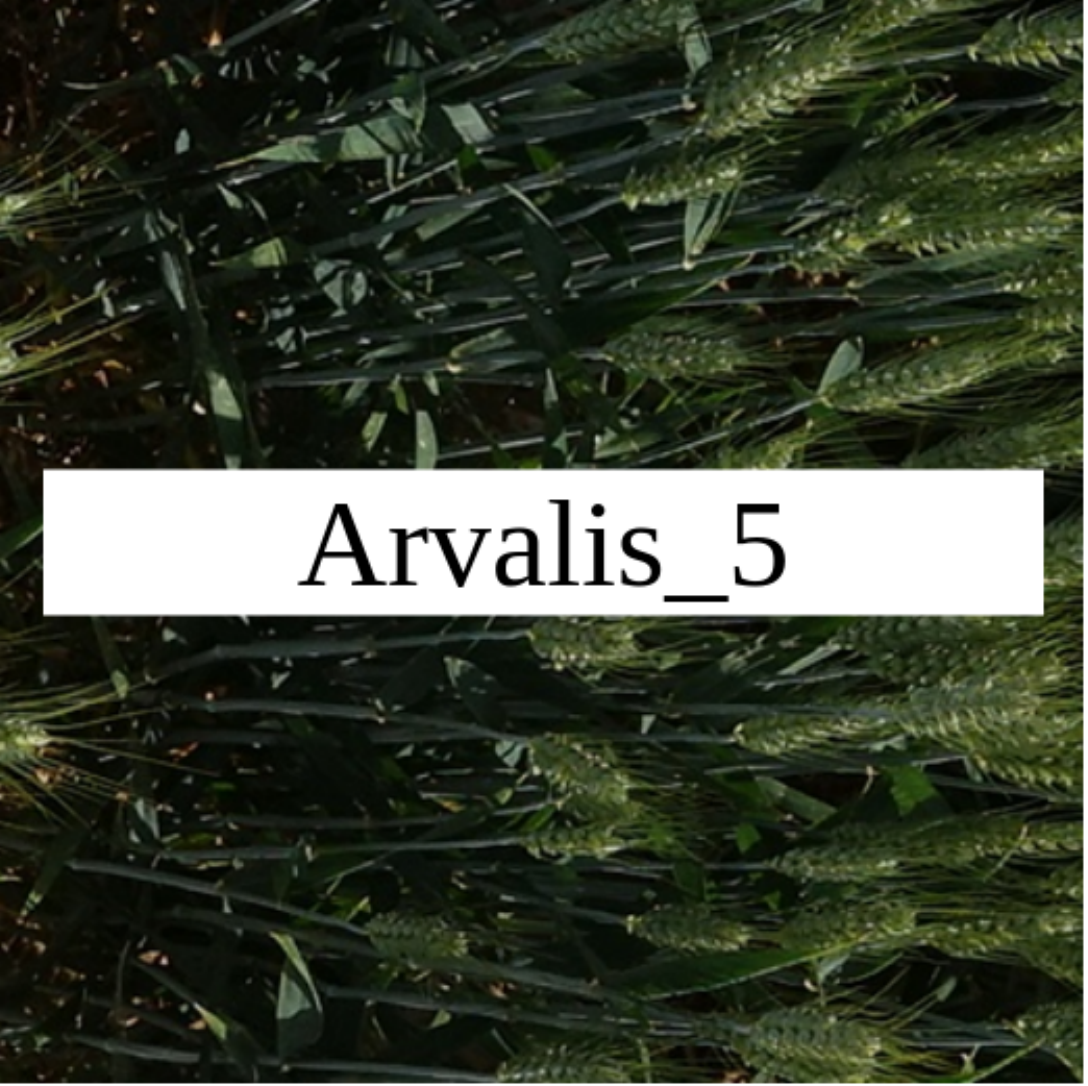}} & $\mathcal{S}$                                                                   & 0.290              \\
                  & $\mathcal{F}_{\eta + \rho_1}$                                                                     & 0.193     \\
                  & $\mathcal{F}_{\eta + \zeta + \rho}$                                                           & 0.271              \\ \hline
\multirow{3}{*}{\includegraphics[width=\CellLength, height=\CellLength]{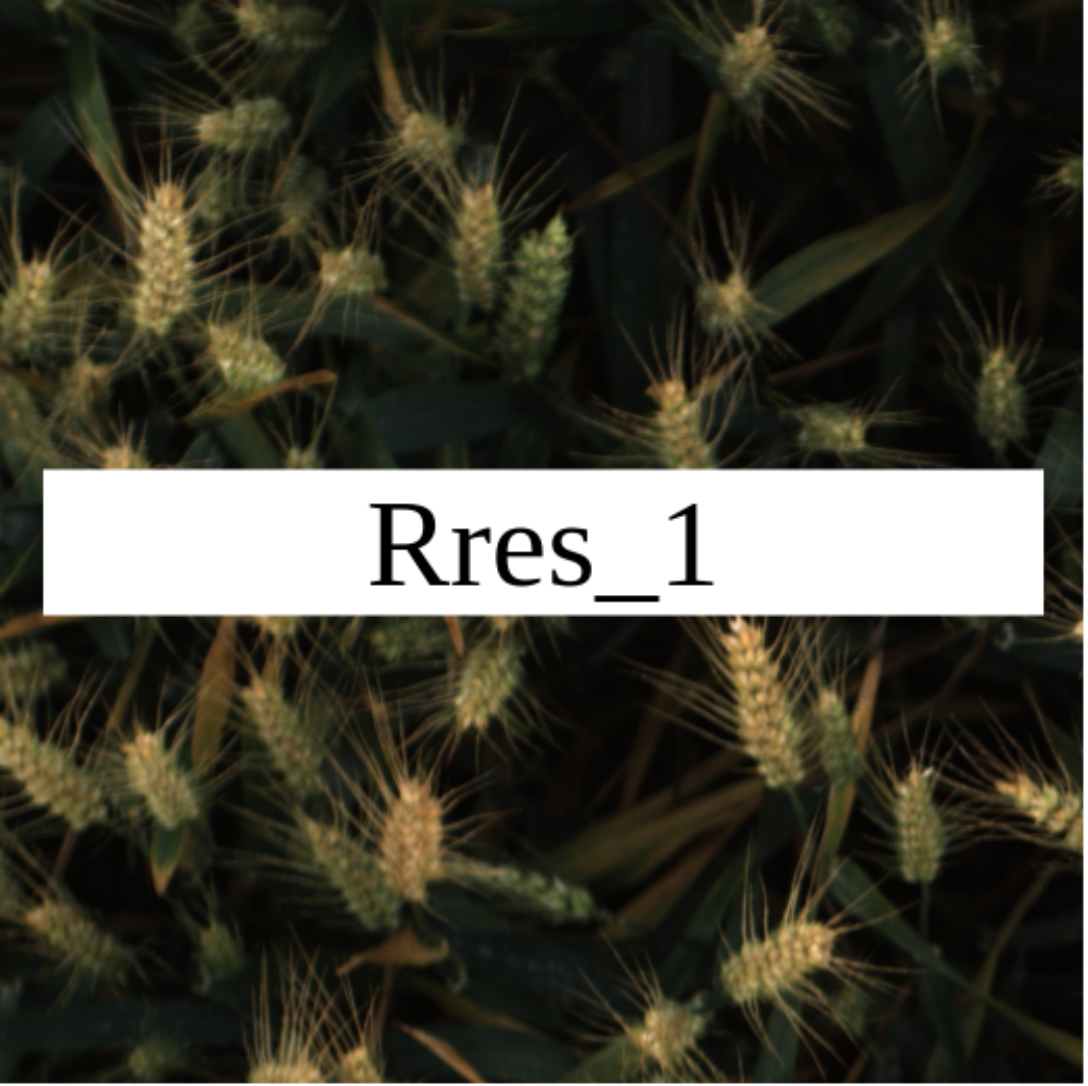}} & $\mathcal{S}$                                                                   & 0.601              \\
                  & $\mathcal{F}_{\eta + \rho_1}$                                                                    & 0.599              \\
                  & $\mathcal{F}_{\eta + \zeta + \rho}$                                                            & 0.769              \\ \hline
\multirow{3}{*}{\includegraphics[width=\CellLength, height=\CellLength]{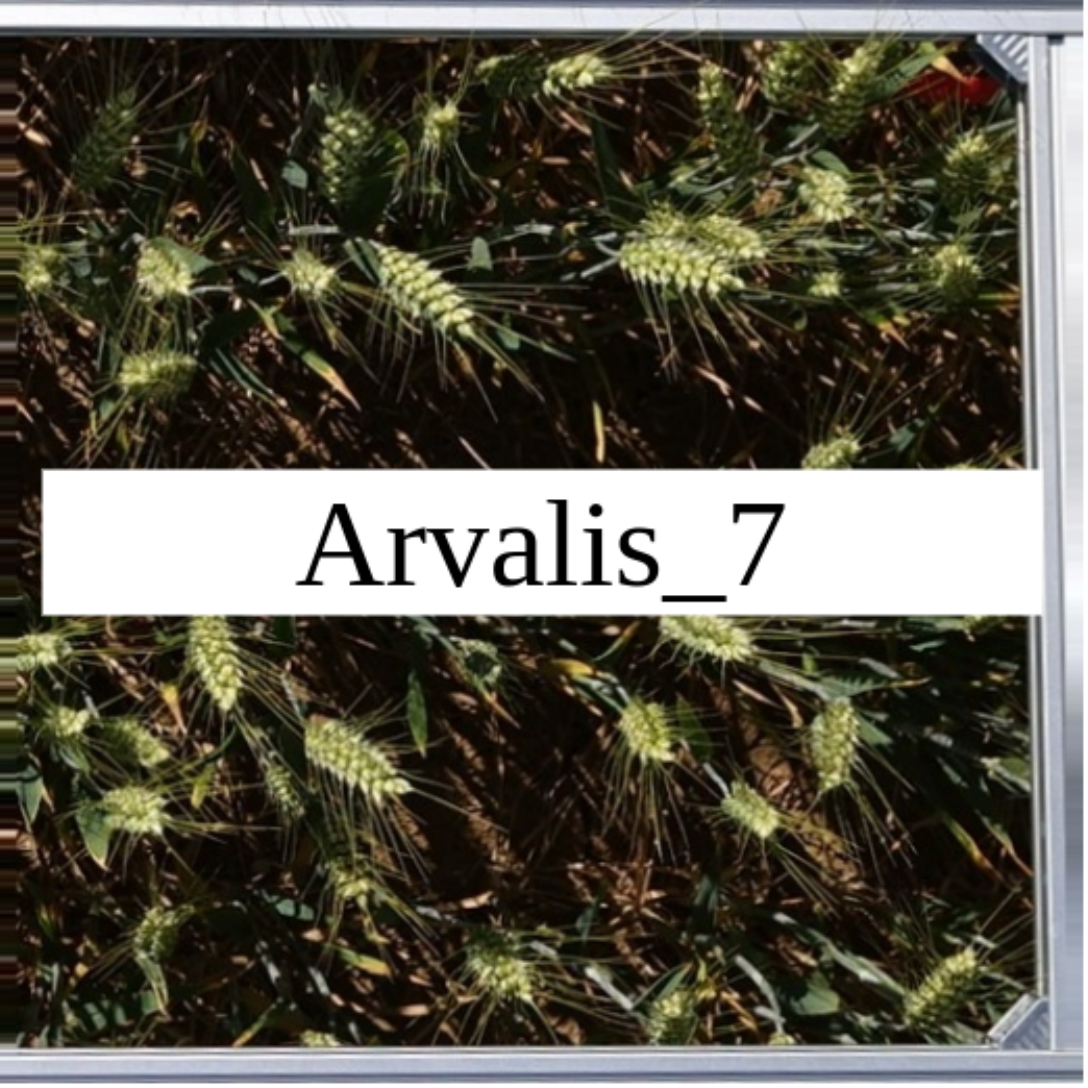}} & $\mathcal{S}$                                                                   & 0.583              \\
                  & $\mathcal{F}_{\eta + \rho_1}$                                                                      & 0.795               \\
                  & $\mathcal{F}_{\eta + \zeta + \rho}$                                                            & 0.748               \\ \hline
\multirow{3}{*}{\includegraphics[width=\CellLength, height=\CellLength]{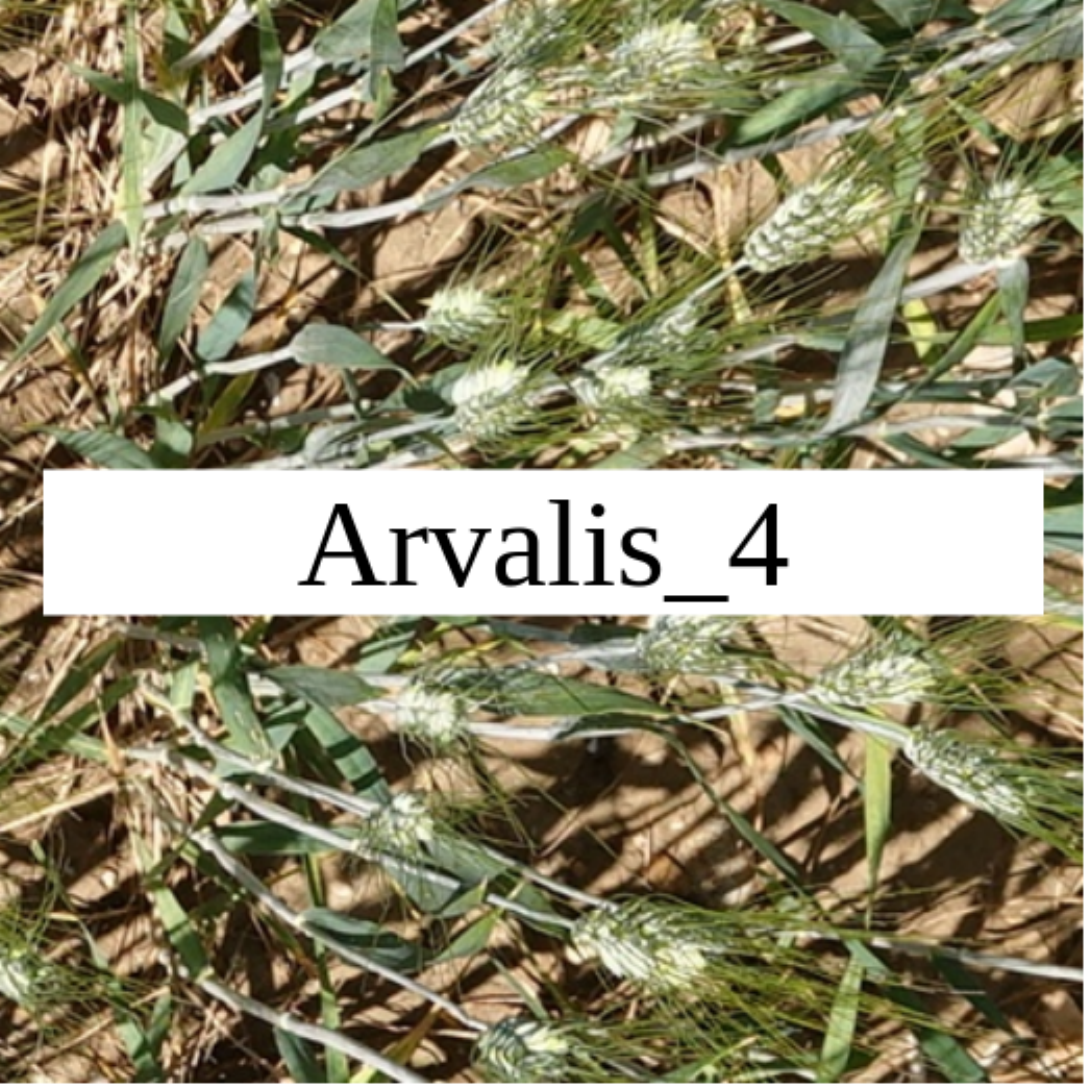}} & $\mathcal{S}$                                                                    & 0.156              \\
                  & $\mathcal{F}_{\eta + \rho_1}$                                                                    & 0.286     \\
                  & $\mathcal{F}_{\eta + \zeta + \rho}$                                                           & 0.356              \\ \hline
\multirow{3}{*}{\includegraphics[width=\CellLength, height=\CellLength]{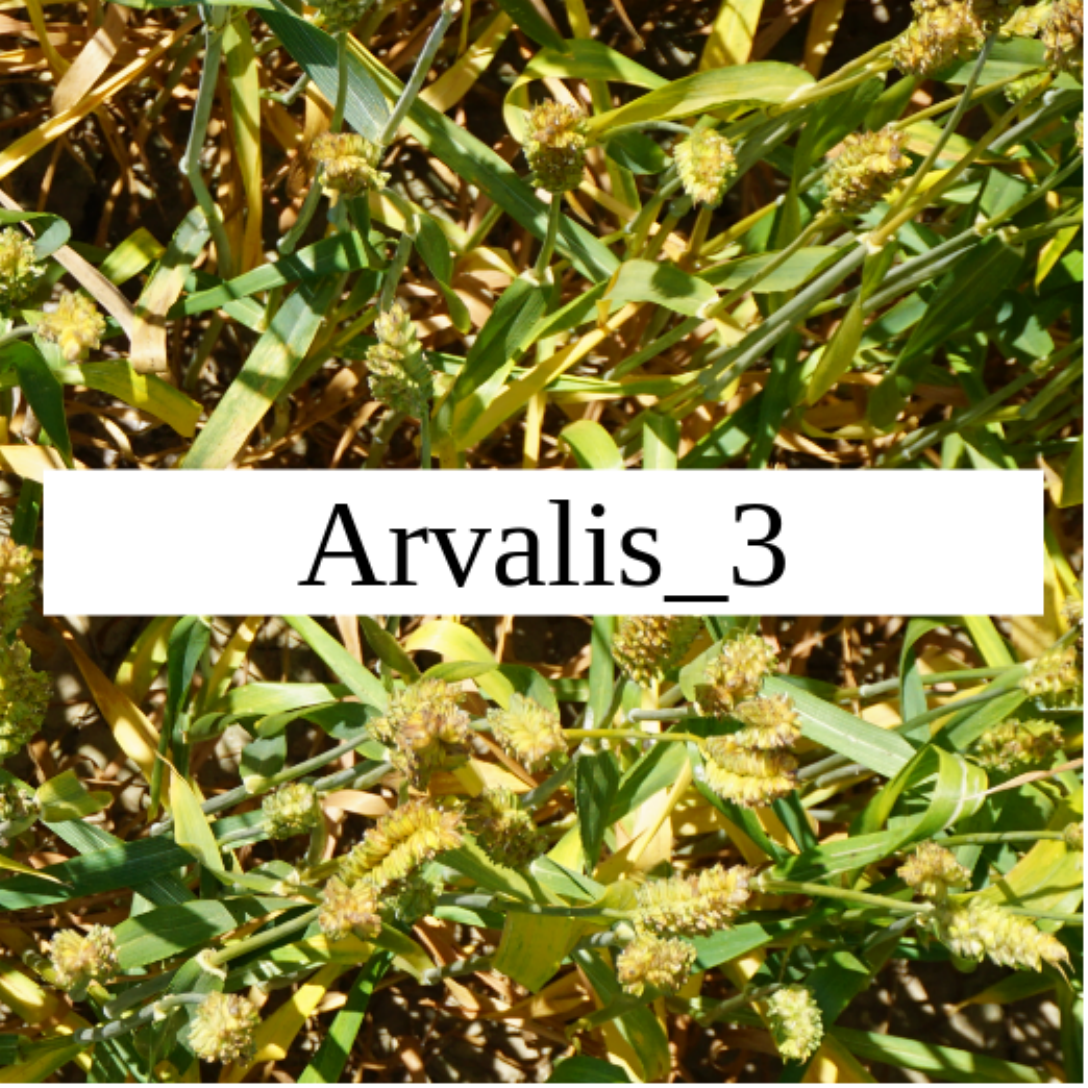}} & $\mathcal{S}$                                                                    & 0.582              \\
                  & $\mathcal{F}_{\eta + \rho_1}$                                                                      & 0.492     \\
                  & $\mathcal{F}_{\eta + \zeta + \rho}$                                                           & 0.667              \\ \hline
\multirow{3}{*}{\includegraphics[width=\CellLength, height=\CellLength]{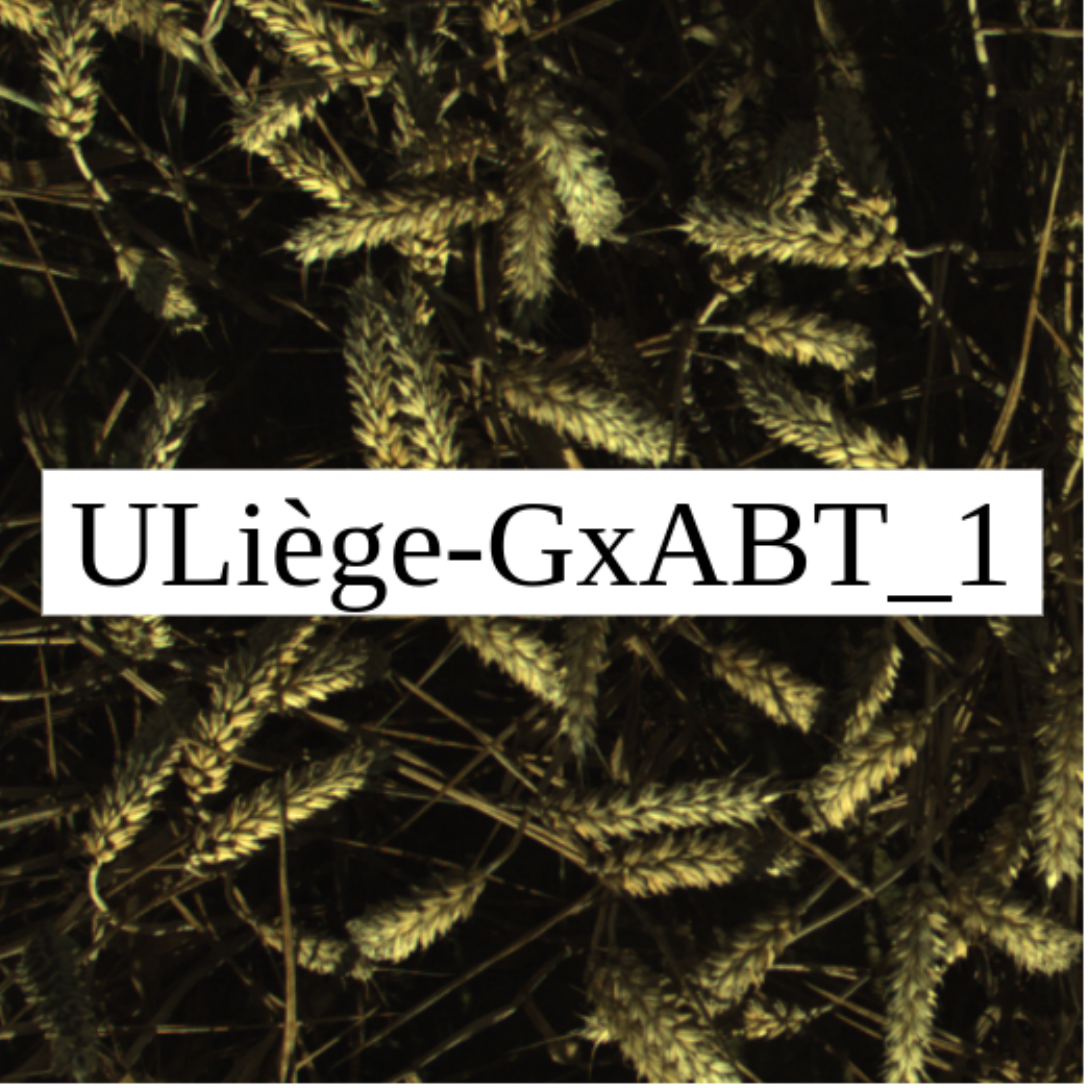}} & $\mathcal{S}$                                                                    & 0.884              \\
                  & $\mathcal{F}_{\eta + \rho_1}$                                                                     & 0.648     \\
                  & $\mathcal{F}_{\eta + \zeta + \rho}$                                                            & 0.805              \\ \hline
\multirow{3}{*}{\includegraphics[width=\CellLength, height=\CellLength]{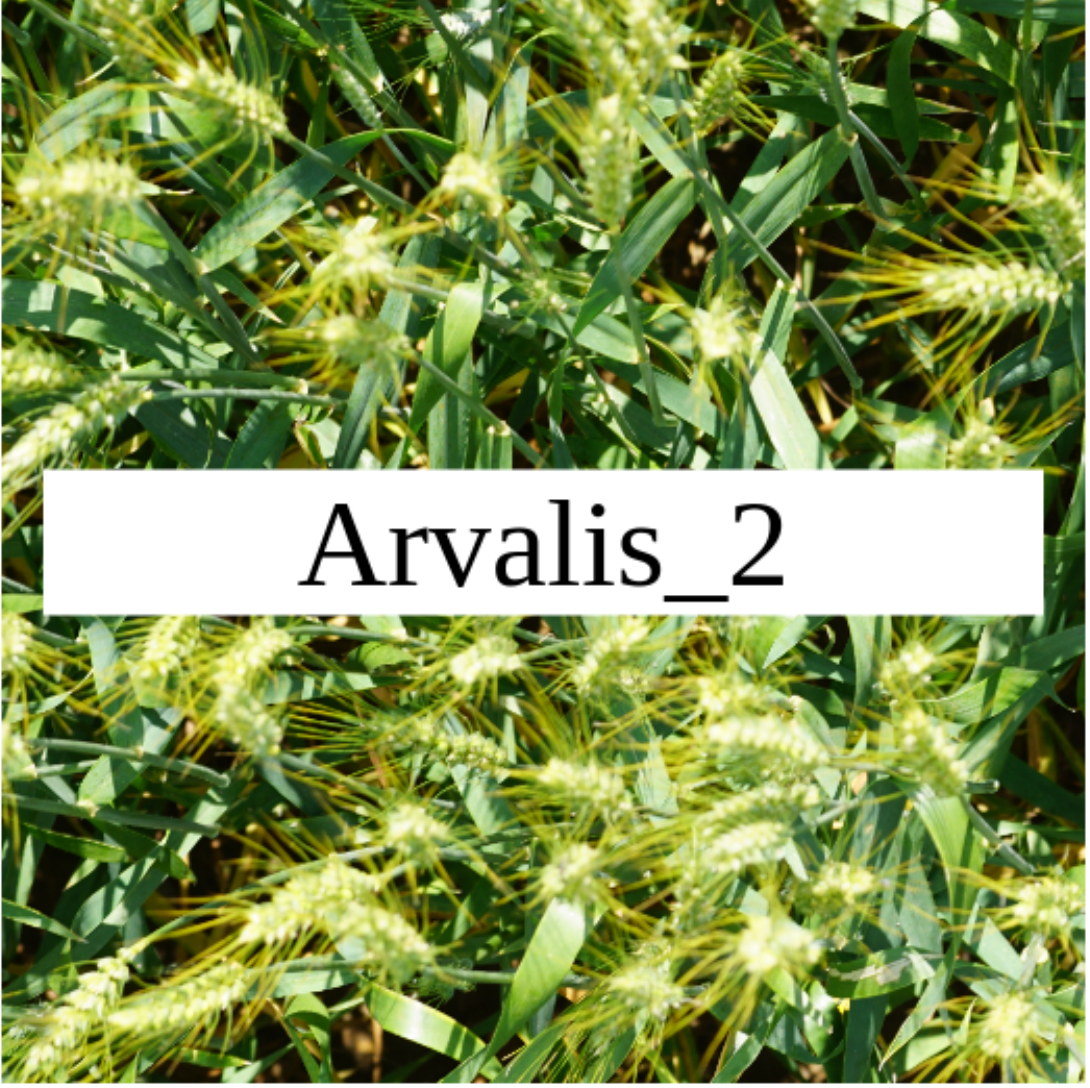}} & $\mathcal{S}$                                                                    & 0.501              \\
                  & $\mathcal{F}_{\eta + \rho_1}$                                                                    & 0.488     \\
                  & $\mathcal{F}_{\eta + \zeta + \rho}$                                                            & 0.686              \\ \hline
\multirow{3}{*}{\includegraphics[width=\CellLength, height=\CellLength]{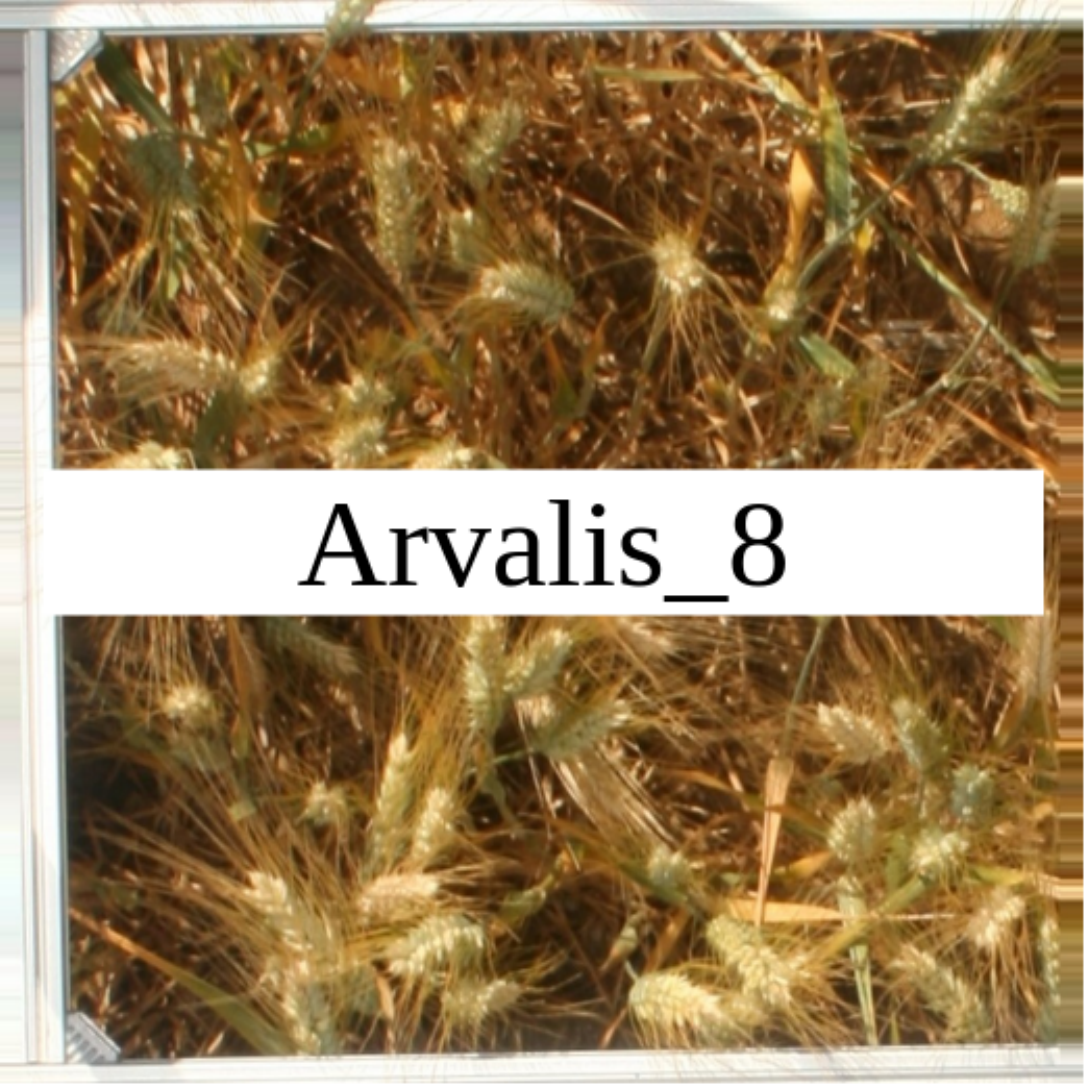}} & $\mathcal{S}$                                                                    & 0.629              \\
                  & $\mathcal{F}_{\eta + \rho_1}$                                                                     & 0.496     \\
                  & $\mathcal{F}_{\eta + \zeta + \rho}$                                                            & 0.520             \\ \hline\hline
\end{tabular}
\end{minipage}
\end{table}

% -------------------
% Discussion 
\section{Discussion}
Precision agriculture aims to integrate advanced technologies to address various challenges in the agricultural sector, enhancing productivity, efficiency, and profitability while minimizing waste and environmental impact. Deep learning approaches play a crucial role in enabling automated decision-making capabilities. By automating the processing of visual data from agricultural fields, these decision-making processes can be significantly improved and scaled up.\par
However, the development of DL-based techniques encounters challenges stemming from the dynamic nature of agricultural fields, characterized by diverse growth stages, inconsistent weather conditions, and variable lighting. As such, models developed based on one snapshot of these fields often are not generalizable across different growth stages or environmental conditions. Furthermore, developing large-scale annotated datasets across various growth stages and environmental conditions is time-consuming and expensive. These challenges hinder the development of generalizable DL-based solutions for various agricultural tasks. Consequently, developing methodologies that allow DL-based solutions to generalize across different conditions (domains) could facilitate the widespread adoption of these technologies in the agricultural sector.\par 
In response to these challenges, we developed a semi-self-supervised domain adaptation technique that employs deep convolutional neural networks and a probabilistic diffusion process. This approach was developed with minimal data annotation, requiring only three manually annotated images. We classify this method as semi-supervised because it leverages a small amount of annotated data (three images) alongside a significant volume of unannotated data. Additionally, this method can be considered self-supervised as it utilizes computationally annotated data for model development.\par
The proposed model demonstrated substantial improvements in performance compared to the baseline model from recent work by Najafian et al.~\cite{najafian2023semi}, while also showing lower variance in model performance across 18 different domains from the GWHD dataset. These 18 domains represent various growth stages of wheat fields and different environmental conditions.\par
In this study, we devised a two-phase model training process. Initially, model $\mathcal{F}_{\eta + {\rho}_1}$ was \deleted{trained}\added{ developed by synthesizing a large number of computationally annotated samples resulted from} \deleted{using} a single sample for \added{image} synthesis and validation, showcasing nearly a $20\%$ improvement over a recent work by Najafian et al.~\cite{najafian2023semi} on the target domain. Subsequently, in the second phase, model $\mathcal{F}_{\eta + \zeta + \rho}$ leveraged additional computationally generated data, further enhancing the performance difference to $28\%$.\par 
As presented in Table~\ref{tab:general_models_performance}, our final model, $\mathcal{F}_{\eta + \zeta + \rho}$, shows $28.1\%$ improvement in Dice score and $25.2\%$ improvement in IoU when compared to model $\mathcal{S}$ from Najafian et al.
The increased performance showcases the utility of the proposed semi-self-supervised domain adaptation technique. \par
In this research, we adopted an extreme strategy by utilizing only two annotated images for training and one for validation. However, we recommend incorporating more manually annotated samples from diverse domains to further enhance the robustness of the proposed approach. Additionally, in this research, we primarily relied on default hyperparameters. Tuning model hyperparameters could lead to further improvements in model performance.
\section{Conclusion}
This research introduces a semi-self-supervised domain adaptation methodology characterized by a dual-stream encoder-decoder model architecture, specifically designed for the downstream task of semantic segmentation of wheat heads. We synthesized a large-scale, computationally annotated dataset from three manually annotated images. We also used unannotated image frames extracted from wheat field videos captured at two different growth stages. The model architecture and training strategy, utilizing both image segmentation and reconstruction, were strategically designed to mitigate challenges associated with domain shift while minimizing the need for extensive data annotation. The external evaluation of our proposed approach on a subset of the GWHD dataset demonstrated a substantial improvement in model performance over a recent work. This underscores the utility of our proposed approach in alleviating domain shift, allowing for the development of generalizable models with minimal manual data annotation, which, in turn, could enable the widespread adoption of DL-based approaches in the agricultural sector.\par

%%%%%%%%%%%%%%%%%%%%%%%%%%%%%%%%%%%%%%%%%%
\section*{Author Contributions}
The methodology of this study was designed and implemented by AG, who also authored the initial draft of the manuscript. Both FM and GHS edited and refined subsequent versions of the paper. FM contributed to the ideation of the research.  FM and GHS provided supervision throughout the study.

\section*{Funding}
This research received no external funding.

\section*{Acknowledgments}
We acknowledge Keyhan Najafian for providing access to the dataset used in this study and for granting access to the baseline model employed for model evaluation.

\bibliographystyle{unsrt}
\bibliography{references}  %%% Uncomment this line and comment out the ``thebibliography'' section below to use the external .bib file (using bibtex) .





% \end{thebibliography}

\end{document}